# Cloth Manipulation Planning on Basis of Mesh Representations with Incomplete Domain Knowledge and Voxel-to-Mesh Estimation

Solvi Arnold*[1], Daisuke Tanaka[2], Kimitoshi Yamazaki[1]

*Abstract*—We consider the problem of open-goal planning for robotic cloth manipulation. Core of our system is a neural network trained as a forward model of cloth behaviour under manipulation, with planning performed through backpropagation. We introduce a neural network-based routine for estimating mesh representations from voxel input, and perform planning in mesh format internally. We address the problem of planning with incomplete domain knowledge by means of an explicit epistemic uncertainty signal. This signal is calculated from prediction divergence between two instances of the forward model network and used to avoid epistemic uncertainty during planning. Finally, we introduce logic for handling restriction of grasp points to a discrete set of candidates, in order to accommodate graspability constraints imposed by robotic hardware. We evaluate the system's mesh estimation, prediction, and planning ability on simulated cloth for sequences of one to three manipulations. Comparative experiments confirm that planning on basis of estimated meshes improves accuracy compared to voxel-based planning, and that epistemic uncertainty avoidance improves performance under conditions of incomplete domain knowledge. Planning time cost is a few seconds. We additionally present qualitative results on robot hardware.

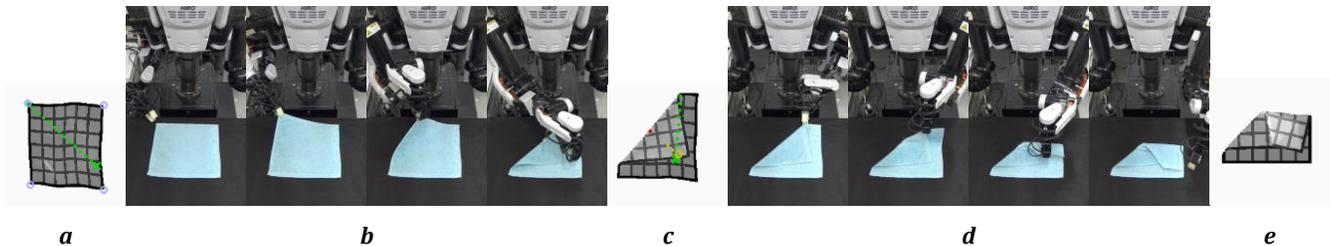

Figure 1. Execution of a two-step manipulation plan. a. Estimate of initial state and first planned manipulation (dotted green line). b. Execution of first manipulation. c. Planned intermediate goal and second planned manipulation (dotted green line). d. Execution of second manipulation. e. Goal state.

## 1. Introduction

An important aspect of the versatility of human intelligence is that we can set goals and flexibly deduce how to achieve those goals. That is, we are capable of solving tasks we have not explicitly learned to solve. Realising such versatility on broad domains is a central challenge to AI. In the present work, we pursue versatile, open-goal planning abilities for the domain of cloth manipulation. Robotic cloth manipulation (CM) is a challenging problem receiving increasing attention in recent years. CM is ubiquitous in household chores, making it an important target for household robotics. However, automation of CM tasks remains rare, even in controlled industrial settings. This low rate of automation is in large part due to the fact that common manipulation methods assume object rigidity, an assumption cloth violates. For example, an assumption of rigidity often allows us to infer a complete object representation from a partial observation of a known object. Also, under assumption of rigidity, it is often relatively straightforward to predict the object's configuration after performing a given manipulation on it from a complete object representation and a specification of the manipulation. Methods built on this assumption do not generalise to cloth. Knowing position and pose of parts of a cloth object constrains but does not determine the position and pose of its other parts, and performing a manipulation on a cloth object will often drastically alter its shape in complex ways, substantially complicating prediction of manipulation outcomes. For highly constrained tasks, we can sometimes avoid the need to deal with these issues. However, if we aim to realise broadly applicable, generalised CM skills in robotic systems, we should address CM's inherent complexities in a structured way.

Humans manipulate cloth intuitively. We plan manipulations for bringing various cloth items into various shape configurations with ease, even for many unseen cases. This suggests a generalised planning ability, although the actual planning process is notoriously hard to verbalise or formalise. We also operate over various levels of precision. When we prepare a shop display we may manipulate on basis of very fine-grained goals and expectations, but when we quickly drape a towel over a rail we may have only a rough expectation of the outcome, and no need for more. The ability to set and pursue goals at various levels of precision and to operate with variable uncertainty levels are likely important factors in the flexibility of our manipulation skills.

---

* Corresponding author: s_arnold@shinshu-u.ac.jp
[1] Department of Mechanical System Engineering, Shinshu University, Nagano, Japan
[2] Department of Science and Technology, Graduate School of Medicine, Science and Technology, Shinshu University, Nagano, Japan

Building on previous work, we pursue manipulation planning abilities for CM with emphasis on flexibility and uncertainty. Specifically, our manipulation planning task assumes open start and goal shapes. While the task is not modelled after any specific household task, the requirements it presents are intended to capture some of the uncertainty and variability that household support robots will have to be able to deal with in order to operate effectively. With regard to operating speed as well, we consider the demands of a household support setting: maximising operating speed is less important than in industrial settings, but the approach should be fast enough to make practical applicability feasible.

We also consider the demands of the household support setting in terms of the robot hardware we assume. Some CM tasks can be solved with relative ease using specialised hardware. However, in a household setting it is preferable to solve a large array of problems with just one or a few pieces of robotic hardware. We assume a dual-handed setup without specialisation to CM, consistent with the common conception of loosely humanoid support robots. We do consider two arms to be the minimum for efficient CM, as many basic manipulations become significantly more complex when performed single-handedly (consider e.g. the basic task of folding a towel neatly in two).

We also address the issue of planning with incomplete domain knowledge. Cloth items' shape configurations have essentially infinite degrees of freedom. Explicitly representing item shapes at a decent level of precision yields high dimensional state spaces. Furthermore, the goal of high flexibility requires a sufficiently broad action repertoire. Consequently we have to operate in a sizable state-action space. Having to sample such a large space exhaustively to generate training data is impractical and wasteful (many state-action pairs produce no state transitions of practical interest). Sampling only state-action space regions of interests is preferable, but necessitates a strategy for planning on basis of incomplete domain knowledge. The present work addresses this challenge for the case of CM, using an explicit measure of epistemic uncertainty w.r.t. manipulation outcomes.

## 2. Related work

Here we discuss existing work related to the present work by task setting or approach. As of yet there is no clear dominant strategy for robotic cloth manipulation. Choice of strategy is informed by a number of factors. We discuss some of the various strategies that have been applied so far, with a focus on flexibility and operation speed.

Fixed routines (Koishihara et al., 2017; Maitin-Shepard et al., 2010) and routines with simple branching (Yuba et al., 2017) have been proposed for specific CM tasks. Such routines have application in industrial settings and for particularly common manipulations, but lack the flexibility to accommodate variable start or goal states. Flexibility requires thinking ahead, which naturally suggests simulation-based solutions (Kita et al., 2014; Li et al., 2015). However, simulation-based approaches face challenges in achieving practical processing times.

### 2.1. Reinforcement learning & forward models

One promising avenue for reconciling flexibility with practical operating speeds is seen in approaches employing neural networks (NNs). NNs have been applied in the context of CM for grasp point detection in a bed-making task (Seita et al., 2018) and prediction of forces exerted on human subjects in a dressing task (Erickson et al., 2018). NNs are also widely used in systems trained through Reinforcement Learning (RL), as have been applied to feedback-based folding (Petrík and Kyrki, 2019), and cloth smoothing (Wu et al., 2020). Various methods using RL to learn from demonstrations have also been proposed over the past years, for e.g. fabric smoothing (Seita et al., 2019) and dynamic manipulation (Jangir et al., 2020).

Our approach employs NNs trained as forward models (FMs). We will refer to such NNs as FMNNs for short. The use of FMNNs for control tasks has been explored on various domains (Ebert et al., 2018; Henaff et al., 2017; Janner et al., 2019; Lesort et al., 2018; Wahlström et al., 2015), albeit in most cases on tasks of lower dimensionality than CM requires. The distinction between planning and model-driven control can be blurry here: both amount to generating actions on basis of the similarity of their predicted outcomes to a given target. However, control tasks typically assume shorter time-steps and fixed goals. Common benchmark tasks are pole-balancing and stable locomotion with a given body plan. The assumption of fixed or open goals, in particular, has consequences for the design of the system around the FM. FMs in the context of control tasks are often embedded in a RL context. A learned FM can replace the task environment in model-free RL to improve sample efficiency (Henaff et al., 2017; Wahlström et al., 2015), or be used as part of a model-based RL algorithm (Clavera et al., 2020). While such strategies are suitable for control tasks, planning as we pursue here requires a higher level of flexibility. RL requires that a reward structure (i.e. an implicit specification of a goal state or conditions to be maintained) is set at training-time. This necessity is at odds with our assumption of variable goals set at run-time. Due to the need to define the reward structure at training time, the vast majority of RL-based work assumes fixed goals. A rare exception is seen in (Jangir et al., 2020), where the policy and reward functions are made conditional on a low-dimensional goal variable, which makes it possible to vary some aspects of the goal at run time. However, if and how this strategy could be extended to address high-dimensional goal shapes as considered in the present paper is unclear.

Here, instead of learning a policy, we use the fact that NNs are differentiable to obtain gradients for the action inputs, allowing for fast action search on basis of a goal given at run-time. We note that the differentiability of FMNNs has also been exploited in various fixed-goal RL approaches (Clavera et al., 2020; Henaff et al., 2017; Pereira et al., 2018). Our approach is similar to that of (Henaff et al., 2017): backpropagation through the FMNN provides gradients for the action inputs, that guide action search toward high-quality actions (although in (Henaff et al., 2017) the FM is subsequently distilled into a faster fixed-goal policy through RL). Also similar, albeit with a non-NN FM, is the approach of (Watter et al., 2015). A second similarity shared with a subset of these systems is the use of latent state representations. Learning FMs on state representations at their full dimensionality can quickly become computationally impractical. Moreover, it can be advantageous to let an encoder element extract the particular features that are important for prediction. This can be done by training encoder (and decoder) networks in end-to-end fashion together with the FMNN itself, as we will do here.

In addition to the assumption of open goals, the task we target also differs substantially from the typical RL tasks in a number of quantitative aspects. We noted the high dimensionality of adequate cloth shape representations. Ours is 3072D, which exceeds typical state dimensionalities in the current RL literature by a large margin. On the other hand, the number of steps we reason over is small, capping out at 3 in the present work where the control literature often considers tens of steps. However, the amount of time covered by a single forward pass through our FM is unusually long: a single pass corresponds to grasping, moving, and releasing the cloth item. Consequently, measured in real-time the amount of lookahead we consider is by no means short. Hence the present work can in part be seen as an exploration of the effectivity of FMNNs in the high-dimensionality, long-timestep, short-rollout regime, which is an important domain for high-level planning.

### 2.2. FORWARD MODELS IN CLOTH MANIPULATION

For CM specifically, a few FM-based approaches have been proposed over the past years. Yang et al. (2017) report application to fine control in a CM task, albeit with fixed goal states. We previously proposed a system for open-goal, multi-step, dual-handed, FM-based CM planning, using voxel representations and latent encodings thereof (Arnold and Yamazaki, 2019a; Tanaka et al., 2018). Kawaharazuka et al. (2019) apply a similar approach to string and cloth manipulation tasks involving momentum, using 2D images as input. Subsequently, Yan et al. (2020) propose a method using a forward model in latent space, introducing contrastive learning to structure the latent space. Cloth states are represented as RGB images, and (single-step) manipulations are planned by randomly sampling actions and picking the actions for which the predicted outcome is closest to the goal. The method is applied to single-handed manipulation of ropes and cloth, with various goal shapes for rope. For complex goal shapes as we consider here, one-step greedy action generation is limiting. Lack of multi-step foresight can lead the system into dead-end states from which the goal cannot be reached without backtracking (which itself requires multi-step foresight). Hoque et al. (2020) propose a method using RGBD state representation and multi-step manipulation planning by CEM (Cross-Entropy Method), and apply it to cloth smoothing and basic single-handed folds.

The present work builds on the approach of (Arnold and Yamazaki, 2019a). Although spatial dimensionality and channel counts vary, all the above methods have in common that they use rasterisations (voxel volumes, images) of the workspace for state representation. These representations all have similar shortcomings. They are limiting in that they do not capture a topological understanding of the cloth's shape configuration. One of our core contributions here is that we perform FM-based manipulation planning on basis of explicitly topological representations (mesh models). For this purpose, our approach incorporates mesh estimation.

### 2.3. MESH ESTIMATION

Obtaining mesh representations of cloth objects is challenging. For mesh-based planning to be of use in robotic CM requires fast and robust mesh-estimation routines. Willimon et al. (2012) and Han et al. (2018) propose estimation routines for deformable objects, but the amount of deformation considered is less than many common CM scenarios require. Sun et al. (2015) generate high quality 2.5D representations of wrinkled cloth in the context of a cloth flattening task, but here too the range of shapes considered is limited (spread-out objects with wrinkles). Kita et al. (2014) and Li et al. (2014) consider more complex shapes, but their active observation strategies involve lifting and rotating the object. This naturally disturbs the object's shape, which makes these approaches unsuitable for our task scenario. We previously proposed an approach that uses an NN to generate a probabilistic mesh estimate on basis of a voxel representation of the object shape, which is subsequently refined using an energy minimisation procedure (Arnold and Yamazaki, 2019b). We integrate a variant of this approach as a subsystem in our planning system.

### 2.4. INCOMPLETE DOMAIN KNOWLEDGE

Another core element of our approach is that we introduce measures to improve robustness under conditions of incomplete domain knowledge. Sampling the full state-action domain to collect training data is costly and wasteful, as many state-action pairs are simply not particularly useful or interesting. This issue is compounded significantly when we consider dual-handed manipulations, due to the increased dimensionality of the action space. Our solution for planning with incomplete domain knowledge employs multiple instances of the FMNN. The use of network ensembles to boost performance or improve robustness has a long history (Granitto et al., 2005; Hansen and Salamon, 1990). Examples of ensemble methods are also found in the RL literature. Janner et al. (2019) and Clavera et al. (2020) use ensembles of FMs to counteract model bias. In these algorithms trajectories for policy training are sampled from a randomly selected model instance from the ensemble. As noted above, our approach does not involve policy training, and consequently our use of multiple FMs differs from the RL setting. We will use a pair of FMs to calculate an explicit, differentiable measure of epistemic uncertainty, and use this measure to guide back-propagation-based plan search.

3. CONTRIBUTIONS

The present work improves on (Arnold and Yamazaki, 2019a) in a number of ways, with the major areas of improvement being cloth representation, integration of a mesh estimation routine, robustness against incomplete domain knowledge, and the accommodation of grasp point detection routines.

3.1. CLOTH REPRESENTATION

In (Arnold and Yamazaki, 2019a), cloth states are represented in voxel format. Voxel (as well as point cloud) representations have two major advantages for CM. First, the representations are comparatively easy to obtain. Second, these representations are invariant to differences in colour and patterning of cloth items, and robust against variable light conditions. Voxel representations additionally have the advantage that they can be processed by conventional neural network architectures directly (although we note the existence of Point Net (Qi et al., 2017) and work building thereon).

The voxel-to-voxel approach, as well as pixels-to-pixels approaches (Henaff et al., 2017; Hoque et al., 2020; Wahlström et al., 2015), are attractive for their conceptual simplicity: no special-purpose, engineered representation formats need to be considered. When latent representations are used, discovering a suitable representation format is left to the training process. In (Arnold & Yamazaki, 2019a), we demonstrated that substantial planning abilities can be realised using just voxel representations and latent representations thereof. However, the inherent ambiguity of a voxel representation limits how much fidelity can be attained. Many topologically different cloth shapes produce similar voxel representations (for example, the voxel representation of a square cloth folded neatly in two can become highly ambiguous as to which side the fold is on). Consequently, voxel-based planning can produce results that are close to the goal state in voxel representation, but these results may diverge topologically from the intended goal. Additionally, for back-propagation-based planning, action gradients obtained by back-propagating the error between a predicted outcome state and a goal state are not always informative if the states are given in image or voxel formats (as noted in (Yan et al., 2020) for the image case).

Planning with mesh representations instead of voxels for the planning system brings topological information into the planning process, which can be expected to improve fidelity. We employ both deterministic and probabilistic mesh representations, with the latter allowing us to capture aleatoric uncertainty in predicted states. This allows the system to pursue goal shapes with high accuracy where possible, and more loosely when outcome uncertainty is high (generating plans for which the goal is merely estimated to be a plausible outcome). Probabilistic meshes represent uncertainty at per-vertex, per-dimension granularity, so precision levels can even vary over different parts of the object. Furthermore, unlike voxel- and image-based state representations, the action gradients produced by mesh representations are guaranteed to be informative, as the error signal corresponds straightforwardly to mesh similarity between goal and prediction.

Advantages of a mesh-based approach are particularly pronounced when we consider task settings where significant occlusion occurs. Mesh representation allows us to leverage shape information from simulation data that is lost to occlusion in image-based state representation. Our mesh estimation system takes voxel input with occlusion present, but is trained to estimate complete meshes. The positions of occluded vertices cannot in general be determined exactly, but the positions of the visible vertices, as well as the occluded vertices' invisibility itself, constrain the range of possible positions for occluded vertices. The system implicitly learns these constraints to locate occluded vertices, and because we let the mesh representation quantify positional uncertainty per vertex, it can quantify the remaining uncertainty in a principled manner. Finally, the fact that goal states are specified in mesh form has the advantage that it allows us to specify target positions for occluded parts, which is not possible in image-based goal specification.

3.2. MESH ESTIMATION

The use of mesh representations in planning necessitates that we obtain these mesh representations somehow. We integrate the approach of (Arnold and Yamazaki, 2019b) into the planning system in order to obtain mesh estimations from cloth states observed in voxel format. Mesh estimation consists of two steps. First we generate an initial probabilistic shape estimate using a second NN architecture. This estimate is then refined through a short optimisation process that incorporates prior knowledge of the cloth topology. We leverage the predictive abilities of the FMNN to resolve ambiguity of voxel representations in the mesh estimation process. If a cloth state is the result of a manipulation generated by the system itself, we use the predicted outcome of the manipulation as additional prior knowledge to disambiguate the present state in the optimisation process. Hence shape information flows both ways between the mesh estimation system and planning system.

3.3. INCOMPLETE DOMAIN KNOWLEDGE

For many task domains, only parts of the state-action space are of practical interest. In a CM setup, many parametrisations of the manipulation representation will correspond to manipulations that make no useful change to many or most cloth shapes. Which parametrisations have useful effects will depend on the current shape of the cloth, so engineering the state-action space upfront to exclude uninteresting cases quickly becomes infeasible. More fundamentally, it is often hard to formalise what is or is not useful. However, for many domains we can collect sets of relevant examples. A collection (dataset) of examples that samples only part of the full domain can be considered to (implicitly and fuzzily) define a "region of interest" on that domain. A versatile planning system must be able to find good actions or action sequences on basis of experience that is restricted to this region of interest. It should avoid uninteresting regions of the space, without requiring explicit definition of which regions are interesting. However, FMNNs are vulnerable to incomplete domain knowledge, for two main reasons. First, for methods that rely on error gradients (such as the method we pursue below), gradients for unknown regions of the state-action space will be unreliable, potentially causing plan search processes to spend time and computational resources chasing meaningless gradients without approaching the region of interest. Secondly, the NN may spuriously predict outcomes similar to the goal state for state-action sequences leading through unknown regions of the domain. This can lead to generation of spurious plans that fool the system into expecting the goal state as outcome. This problem is closely related to the problem of adversarial examples (Goodfellow & Szegedy, 2014; Szegedy et al., 2013) in NN-based vision. In image recognition, input spaces (e.g. the space of all possible images of a given resolution) are typically so vast that training data can only sample a limited subspace of it (e.g. a set of real-world images). In a typical experimental setup, such NNs are trained and tested on image sets that are assumed to sample roughly the same parts of the input space. This can produce impressive results. However, such NNs can and do produce unexpected classifications for unfamiliar inputs. In the field of image recognition, inputs intentionally designed to elicit erroneous classifications are known as "adversarial examples" (Szegedy et al., 2013), and are an active field of study. In neural network-based planning, training with datasets that only partially sample the input space can similarly lead to the existence of adversarial examples, in the form of what we might call "adversarial plans", as we demonstrate in (Arnold and Yamazaki, 2019c).

The existence of adversarial plans can be a significant problem for these methods, and in some sense more so than in image recognition. Image recognition NNs typically have no control over the input they receive. Adversarial examples in image recognition are the product of "adversarial attacks": attempts by an outside force to deceive the system. In a "white box" attack[1], the attacker may for example use a gradient descent approach to search the input space for images that deceive the NN into producing an erroneous classification (Athalye et al., 2018; Goodfellow and Szegedy, 2014; Kurakin et al., 2016). Now consider NN-based planning. Many of these systems work by actively searching the NN's input space for inputs that minimise a given loss. This means that if adversarial plans exist within the plan space, they appear as additional local optima, and plan search will pursue them just like it pursues valid plans. In other words, NN-based planning essentially implements its own white box attack. We refer to the unintentional generation of adversarial plans as "self-deception". Because the search process will approach adversarial plans just as it approaches valid plans, adversarial plans need not even be common in the plan space to pose a problem.

Ensuring that the NN is trained on complete domain knowledge avoids the self-deception pitfall, but a domain broad enough to produce a versatile system will often be too large for exhaustive sampling to be practical, especially if data is to be sampled on robot hardware. This makes planning on basis of incomplete domain knowledge a problem of significant practical importance. In the present paper we aim to realise versatile CM planning on basis of incomplete domain knowledge, using a dual-network approach for explicit epistemic uncertainty avoidance.

### 3.4. ACCOMODATION OF GRASP POINT DETECTION ROUTINES

A complication in constructing flexible planners for robotic systems is that reality imposes complex restrictions on which actions are physically possible or effective. In the case of CM, we need to account for practical graspability of the cloth item to be manipulated. As noted above we assume non-specialised robot hardware. We assume that the robot is equipped with basic grippers, with finger tips narrow enough to slide under a piece of cloth laid out on a flat surface. Under these assumptions, cloth can be grasped by exposed corners, edges, and fold lines, but it is hard to grip the cloth reliably on smooth parts of its surface. Hence we will want to constrain the set of grasp points to consider for a given state to a set of reliably graspable points. Grasp point detection has its own set of challenges, especially for deformable objects. Graspability will in general depend on the specifics of the object and the robotic system being targeted. We assume a simple candidate grasp point detection routine here, which in a practical application could be switched out for more advanced candidate detection methods suited to specific tasks and robotic hardware.

On a more abstract level, the assumption of constrained grasp points gives rise to a problem setting where each state presents with a variably sized set of continuous values for (part of) the action input (variably sized because the number of grasp point candidates varies with the cloth shape). This type of action space is rarely considered. The action domains of common RL benchmark tasks consist of a fixed discrete set of actions (e.g. moves in grid mazes), or a set of continuous action values (e.g. inverted pendulum balancing). Uncommonly, we see compound action domains consisting of discrete and continuous components (Henaff et al., 2017). Our case does not fit into any of these categories, but this type of action domain occurs naturally in tasks in which robots manipulate complex environments. Addressing this type of input regime in an FMNN-based planning system broadens applicability in practical robotic tasks.

Assuming variably sized sets of real-valued action inputs introduces new complications in the planning system, some linked to the issue of incomplete domain knowledge discussed above. We present grasp point inputs as continuous values to our FMNN, but this leads to a large part of the network's input domain remaining unsampled (state-action pairs with a grasp points outside the candidate set for the state are valid as NN input, but never seen during training). Hence for handling this type of input domain, consideration of epistemic uncertainty is expected to be beneficial. Furthermore, planning involves large numbers of rollouts of the forward model. Having to perform grasp point detection on each subsequent state prediction would be computationally impractical (especially if we consider that grasp point detection may itself be expensive in many cases) and would be complicated by the fact that predicted states are probabilistic. Consequently, the system has to implicitly guess where the cloth will be graspable in future timesteps when generating its manipulation plans.

---

[1] White box attack: An attack where the adversary has full access to the NN.

4. SYSTEM ARCHITECTURE

In this section we first give a global overview of the structure of the system, followed by detailed explanations of its constituent parts.

4.1. OVERVIEW

The core of the system is a neural network trained as a forward model (FMNN) of the behaviour of a cloth item under manipulation. Given a cloth state and a manipulation, the FMNN predicts the resulting post-manipulation state. Manipulations are represented at coarse temporal granularity. A single manipulation grasps the cloth at one or two points, moves these points over a given distance, releases the cloth, and then waits for the cloth to settle in a stable state. This coarse granularity allows us to cover temporally extended manipulation sequences with a limited number of passes through the forward model.

The forward model operates on latent representations that encode mesh representations of cloth states. The mesh representations are obtained using a mesh estimation routine taking voxel representations as input. The estimation routine consists of a voxel-to-mesh network (VtM net) that estimates a probabilistic mesh representation, and an optimisation process ("refinement") that combines this probabilistic mesh representation with prior knowledge of the cloth topology and (if available) the preceding prediction of the current state, in order to generate a plausible deterministic mesh.

Plan generation takes the present state of the cloth and the intended goal state as input (both in mesh format). We apply gradient descent to find the manipulation inputs that should produce the latter from the former, using manipulation input gradients we obtain by backpropagation through the FMNN. Generation of multi-step plans (i.e. manipulation sequences) is achieved by recurrently chaining the FMNN in this process.

Training the system involves two separate training processes. The first trains the FMNN, along with encoder and decoder modules for mapping full-size mesh representations to latent representations and vice versa. The second training process trains the VtM network to estimate probabilistic meshes from voxel representations. Both training processes use the same dataset, which is generated using cloth simulation. Data generation simulates the scenario of performing sequences of random manipulations on a square cloth the size of a hand towel, on a flat work surface.

4.2. REPRESENTATIONS

*4.2.1. Cloth shape & position*

We use three cloth shape representations in total: voxel representation, deterministic mesh representation, and probabilistic mesh representation. We abbreviate the latter two as DMR and PMR, respectively. Voxel representations are binary matrices of resolution 32×32×16. DMRs are real-valued matrices of size 32×32×3. Assuming a 32×32 mesh topology, this matrix assigns (x, y, z) coordinates to each vertex. PMRs resemble their deterministic counterpart, but represent each vertex coordinate as a multivariate normal distribution. They are real-valued matrices of size 32×32×6, with each 1×1×6 subvolume containing a set of means ($\mu_x$, $\mu_y$, $\mu_z$) and standard deviations ($\sigma_x$, $\sigma_y$, $\sigma_z$) probabilistically describing the position of a single vertex in the mesh. The σ values define the diagonal of the multivariate normal distribution's covariance matrix. A more complete probabilistic representations would specify the full covariance matrix, but we wish to constrain the number of outputs to be learned. Example visualisations of a DMR, PMR, and a voxel representation are given in Figures 2, 3, and 4.

Our baseline experiments include a purely voxel-based version of the system. This variant features probabilistic voxel representations: real-valued matrices of resolution 32×32×16 with each value indicating the probability of the corresponding voxel being occupied.

Cloth shape representations are always centred in the XY plane. We find the cloth centre point by projecting the shape onto the XY plane to obtain a 2D image, and calculating the coordinate averages over all pixels corresponding to the cloth. The centre point coordinates are stored separately as a representation of the cloth state's position, and the cloth is then shifted in the XY plane to bring the centre point to (0,0). By splitting states into a shape component and a position component, we obtain position-invariant shape representations, allowing generalisation over positions. Position information can, however, be important. We may want to plan under restriction of a limited work surface or motion range, for example. In the present work, we do not consider such constraints, but we note that by adding the predicted offsets to the predicted vertex coordinates, we can recover a mesh prediction that is located in the workspace, and define additional planning losses thereon as suits a given scenario.

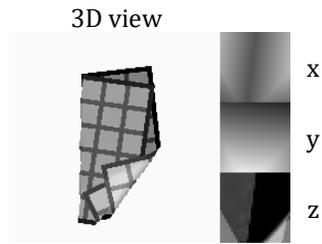

Figure 2. Example of a deterministic mesh representation (DMR). Left: Top-down 3D rendering of the mesh. Right: x, y, and z coordinates mapped as colour gradients in uv-space (e.g. the panel marked x visualises the 32×32 matrix defining the x-coordinates of the mesh as grey values). Grey values for z-coordinates are boosted for visibility. The grid texture on the 3D rendering is added for visualisation purpose only. The texture is not present in cloth observations and does not correspond to the mesh resolution (32×32).

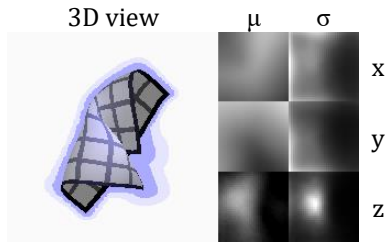

Figure 3. Example of a probabilistic mesh representation (PMR). Left: Top-down 3D rendering of the μ-component of the mesh. Lilac shading around the cloth indicates the $\sigma_x$ and $\sigma_y$ components, and lilac cast on the cloth indicates the $\sigma_z$ component. Right: $\mu_x, \mu_y, \mu_z, \sigma_x, \sigma_y, \sigma_z$ components mapped as colour gradients in uv-space (geodesic space). Grey values for $\mu_z$ and $\sigma_z$ are boosted for visibility.

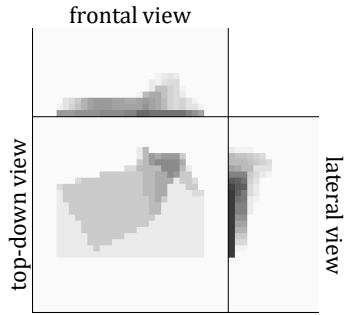

Figure 4. Example of a voxel representation. Generated from simulation data with no noise added. Each panel is generated by taking the mean voxel value over each voxel column parallel to the viewing angle, with darker shades indicating larger proportions of 1-voxels in the column.

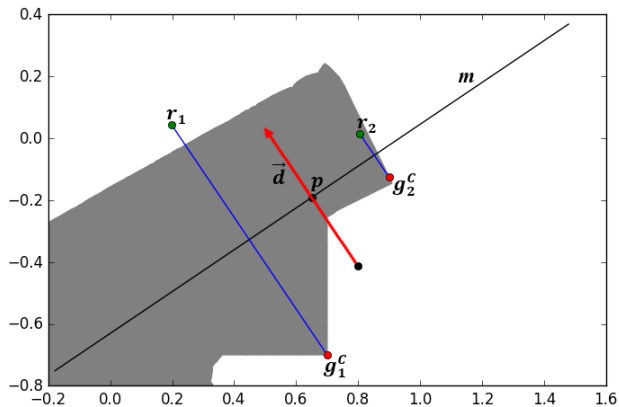

Figure 5. Example calculation of trajectories (blue lines) and release points $(r_1, r_2)$ from grasp points $(g_1^C, g_2^C)$ and displacement vector $\vec{d}$. The grey shape in the background is the silhouette (2D projection) of the cloth.

*4.2.2. Manipulation*

Manipulations are represented as real-valued vectors of length six. The first four values define two grasp points, $g_1^G$ and $g_2^G$, using geodesic coordinates ($u$, $v$), with the cloth running from (-1, -1) to (1, 1). The second grasp point can take null values, indicating a single-handed manipulation. The last two values in the vector are a displacement vector $\vec{d}$ given in 2D Cartesian coordinates ($x$, $y$).

Given a cloth state, this representation determines grasp point trajectories as follows. Cloth states (meshes) map geodesic space to Cartesian space. We let the given cloth state map geodesic grasp points $g_i^G$ to Cartesian grasp points $g_i^C$. For two-handed grasps, we then compute point $p$:

$$p = \frac{g_1^C + g_2^C}{2} + \frac{\vec{d}}{2} = \frac{g_1^C + g_2^C + \vec{d}}{2} \qquad (1)$$

Let $m$ be a line through $p$ perpendicular to $\vec{d}$. The $x$ and $y$ coordinates for Cartesian release points $r_i$ are found by mirroring $g_i^C$ over line $m$ on the XY-plane. Figure 5 shows an example. For cases where the angle between $m$ and the line defined by $(g_1^C, g_2^C)$ is very acute, $g_1^C$ and $g_2^C$ may fall on different sides of $m$. These cases are excluded in data generation (but could in principle occur in plan generation). For single-handed grasps, the $x$ and $y$ coordinate of the single release point $r_1$ is given by $g_1^C + \vec{d}$.

The z coordinate for $r_i$ is given by $g_i^C.z + min(0.2, k/2)$, where $g_i^C.z$ is the z-coordinate of point $g_i^C$ and $k$ is the distance between $g_i^C$ and $r_i$ in the XY-plane. We find the unique circle $c$ centred at height $g_i^C.z$, perpendicular to the XY plane, and passing through $g_i^C$ and $r_i$. The shortest arc segment of $c$ connecting $g_i^C$ and $r_i$ defines the trajectory for point $i$.

The manipulation format improves on that of (Arnold and Yamazaki, 2019a) in terms of flexibility and similarity to human cloth manipulation. In (Arnold and Yamazaki, 2019a), dual-handed manipulations used the same trajectory for both points, translated to be rooted at the two grasp points. When the angle between the displacement vector and the line through the grasp points deviates strongly from 90°, this format often produces manipulations that cause the cloth item to slide over the work surface and/or crumple. These manipulations have little practical use, and human manipulators do not seem to employ them much. Manipulations that cause the cloth to slide are also hard to generalise from to different cloths or work surfaces (or from simulation to reality), as sliding behaviour is quite sensitive to the friction between cloth and work surface. Trajectories produced by the new format rarely cause the cloth to slide over the work surface, and better resemble human cloth manipulation. Consider for example the manipulation that folds a flaring skirt in two over its vertical axis after grasping it at the waistband and lower edge. The grasp points will lie at different distances from the fold line ($m$), and the grasp points will be moved by different distances in the same direction in the XY plane. The new format extends the manipulation repertoire to include such manipulations, while excluding the abovementioned manipulations of low practical value. We note that the new format produces trajectories resembling those considered in (Van den Berg et al., 2010), although our motion in the z dimension is rounded instead of triangular.

An even more flexible format would be to let each hand follow a separately defined trajectory. However, it is hard to isolate which of such manipulations meaningfully add to the repertoire. Many instances can be covered by simultaneous execution of single-handed manipulations, and many instances would result in tearing of the cloth, which introduces its own set of complications.

4.3. GRASP POINT DETECTION

Grasp point detection is challenging problem under active research (Li et al., 2019). In CM, there are often restrictions on where a given robot can reliably grasp the cloth, and again not all points will be of equal interest for grasping. To constrain grasp points to actually graspable points of potential interest, we introduce a simple routine detection routine. This routine plays a role in both data generation and planning. We first projects the cloth mesh onto the XY plane to obtain its top-down silhouette. We then use Shi &Tomasi's corner detection algorithm (Shi and Tomasi, 1994) to obtain the corner points of the silhouette, and discard concave corners. The resulting set of convex corners comprises the graspable point set for the cloth shape. These points will generally be easy to grasp reliably with a wide variety of robot hands. Different detection routines could be plugged in here to suit different robot platforms or tasks, as the larger system does not depend on the details of the routine.

As noted above, grasp points are represented as geodesic ($u$, $v$) coordinates in the manipulation definition. We convert grasp points from cartesian to geodesic space by assigning the ($u$, $v$) coordinates of the vertex nearest to the grasp point in Cartesian space.

4.4. DATA COLLECTION

We collect cloth manipulation examples in simulation, using the ARCSim cloth simulator (Narain et al., 2012; Narain et al., 2013). We construct a scene containing a square cloth laid out flat on a level work surface. Manipulation sequences of length three are applied to this cloth, after which the sequence of manipulations and resulting cloth states are stored to the dataset. We generate single and dual-handed manipulation examples in a ratio of 1:2 (a larger proportion of dual-handed manipulations is generated because the domain of dual-handed manipulations is significantly larger). Manipulation values are randomly generated under the following restrictions.

1. Graspability restriction. Grasp points are restricted to the set of points generated by the detection routine described above. Assuming the routine accurately captures the grasping abilities of the target platform, this restriction allows us to constrain data collection to examples that are pertinent to the platform.

2. Fold restriction. Manipulations that have little or no effect on the cloth shape are of limited interest. To avoid (a large proportion of) such cases we restrict displacement vectors as follows. First we compute a reference point $q$ on basis of the manipulation. For one-handed manipulations, $q = g_1^C + 0.8\vec{d}$. For dual-handed manipulations, $q = (g_1^C + g_2^C)/2 + 0.8\vec{d}$. Only manipulations for which $q$ lies on the cloth (i.e. falls within the projection of the cloth on the XY plane) are included in the dataset. Preliminary experimentation indicated that the majority of manipulations that meet this condition produce folds.

3. Displacement distance restriction. Manipulations with very short displacement distances tends to make no appreciable change to the cloth shape, as very short folds will often undo themselves during shape stabilisation. We enforce that the length of displacement vector $\vec{d}$ should be at least 0.35 times the side of the length of the cloth. We let actions with displacement vectors lengths below this threshold represent null-manipulations instead (null-manipulations are not generated for the dataset, but are derived from valid manipulations in our data augmentation procedure (see Section 4.6), and can be generated by the system as part of a manipulation plan).

These restrictions help to produce a dataset that contains a higher proportion of interesting samples, with fewer examples being overly crumpled and fewer examples just being simple position shifts. We believe that data obtained under these restrictions is more representative of human cloth folding behaviour than unrestricted sampling, and allows us to zoom in on manipulations of higher practical value. However, data collected under these restrictions amounts to incomplete domain knowledge, necessitating countermeasures.

The ARCSim simulator employs adaptive remeshing, changing the mesh topology dynamically to best express the shapes of cloth items as they deform. Our system assumes a fixed 32×32 "grid" topology. While adaptive remeshing can be disabled, this negatively affects the realism of the deformation dynamics as well as shape stability. We instead chose to interpolate regularised meshes from the relevant simulation frames for system input.

We generate a dataset of 3693 sequences, which we split into training, test, and validation sets of 3493, 100, and 100 sequences, respectively.

4.5. PROBABILISTIC EM*D NET (pEM*D)

This NN, extending the architecture proposed by (Arnold and Yamazaki, 2019a), consists of an encoder module E, a manipulation module M (the FMNN), and a decoder module D, with functions defined as follows. Encoder net E maps DMR $s_i$ to its latent representation $\hat{c}_i$.

$$E(s_i) = \hat{c}_i \tag{2}$$

Latent representations are real-valued vectors of length 512. Manipulation net M maps a tuple consisting of latent representation $c_i$, memory trace $t_i$, and manipulation $m_i$ to a latent representation $\hat{c}_{i+1}$ of the predicted outcome shape, new memory trace $t_{i+1}$, and a prediction of centre point shift $\Delta\hat{a}_{i+1}$.

$$M(\hat{c}_i, t_i, m_i') = (\hat{c}_{i+1}, t_{i+1}, \Delta\hat{a}_{i+1}) \tag{3}$$

Where $m_i'$ is $m_i$ with its grasp points shifted by $-a_i$, with $a_i$ denoting the centre point for $s_i$. As centre point movement is relative to the pre-manipulation cloth state, no centre point location needs to be included in M's input. Memory traces, too, are real-valued vectors of length 512. We let $t_0$ be the zero vector. Decoder net D maps a latent representation $\hat{c}_i$ of a cloth state to its PMR $\hat{s}_i$.

$$D(\hat{c}_i) = \hat{s}_i \tag{4}$$

We expect the use of probabilistic output to improve planning performance. During plan generation (detailed below), the predicted output for numerous plans is compared against the goal state. The presence of the σ component in the prediction (along with a planning loss taking it into account) allows planning to weigh differences between goal and prediction in accordance with the prediction's confidence, at per-vertex-coordinate granularity.

By propagating through M recurrently (with new manipulation inputs at every pass), we can generate predictions for multi-step plans. For notational convenience we define the following shorthand:

$$EM^nD(s_i, m) = D(\hat{c}_{i+n}) \tag{5}$$
$$\hat{c}_{i+1} = \pi_1\big(M(\hat{c}_i, t_i, m_i')\big)$$

Where $m$ is a manipulation sequence of length $n$, and $\pi_i$ is the $i$th projection (i.e. $\pi_i$ selects the $i$th element of its argument tuple).

Size parameters of the pEM*D net are given in Table 1, and its structure is illustrated in Figure 6. Connectivity in the M module mixes regular propagation, residual propagation, and skip connections. Propagation from the first 512 neurons in layer $i$ to the first 512 neurons in layer $i+1$ is residual (i.e. values from the first 512 neurons are copied to the next layer). Each layer $i$ for which $i$ is even and *0 < i < 14* also has skip connections projecting into the first 512 neurons of layer *i+2*. This heterogenous connectivity is based on the idea of letting M iteratively transform a latent state representation passing through its residually connected channel. This connectivity pattern performed slightly better than homogeneous connectivity patterns in preliminary experiments, but we have not attempted to verify that the residual channel is in fact used as envisioned. Hidden layers use the hyperbolic tangent activation function. In the decoder module, output neurons representing $\mu$ components use the identity function, and output neurons representing $\sigma$ components use:

$$act_o = e^{act_i} + 0.01 \tag{6}$$

Where *act$_i$* is the incoming activation and *act$_o$* is the outgoing activation. This function has a minimum output value of 0.01, which helps stabilise training, as likelihood loss can fluctuate dramatically when $\sigma$ grows too small. With regards to its probabilistic output, the pEM*D net follows Bishop's Mixture Density Networks (Bishop, 1994), although we use only one distribution per variable.

Given the centre point of the initial state, we can find predicted cloth locations by adding the $\Delta \hat{a}_i$ output of subsequent iterations through the manipulation network.

$$\hat{a}_i = a_0 + \sum_{j=1}^{n} \Delta \hat{a}_j \tag{7}$$

As noted above, values $\hat{a}_i$ can be used to define additional planning losses to constrain plan search to the motion range of the robot, or other such restrictions, although we do not explore this possibility here.

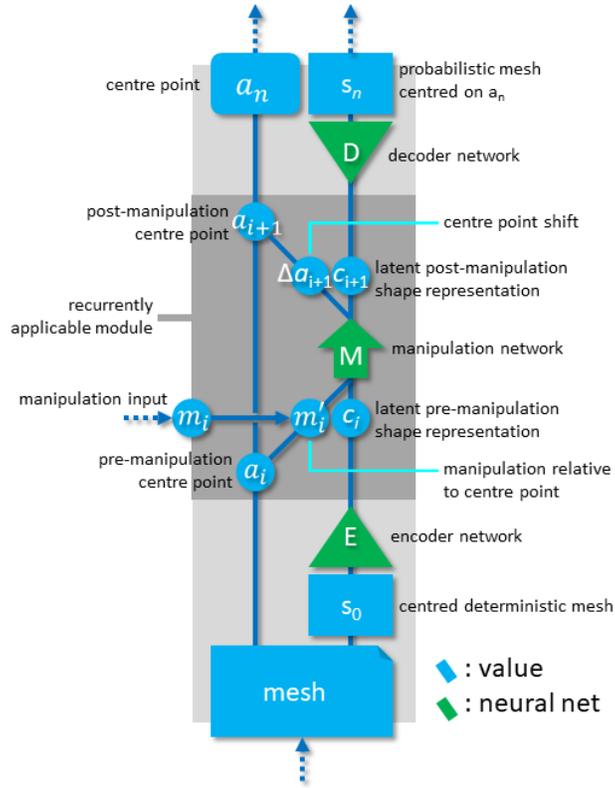

Figure 6. Global structure of the pEM*D net.

TABLE 1. pEM*D NETWORK ARCHITECTURE PARAMETERS

| Module | E | M | D |
|---|---|---|---|
| Input | 32×32×3 ($s_i$) | 512 ($\hat{c}_i$) + 512 ($t_i$) + 6 ($m'_i$) | 512 ($\hat{c}_{i+n}$) |
| Neuron layers | 4 | 16 | 4 |
| Hidden layer sizes | 2048, 1024 | 1024 + 6 ($m'_i$) | 1024, 2048 |
| Output | 512 ($\hat{c}_i$) | 512 ($\hat{c}_{i+1}$) + 512 ($t_{i+1}$) + 2 ($\Delta \hat{a}_{i+1}$) | 32×32×6 ($\hat{s}_{i+n}$) |

## 4.6. pEM*D NET TRAINING

We train modules E, M, and D in end-to-end fashion, thereby forcing E and D to learn a latent representation that facilitates application of manipulations by the manipulation network (instead of simply learning a compact latent representation as would happen if we trained E and D separate from M). The compound net is trained on batches of 64 manipulation sequences of random length between 1 and 3 steps. We use two losses, one measuring shape prediction accuracy ($loss_s$) and one measuring accuracy of the predicted position change ($loss_p$). $Loss_s$ for a single example is defined as follows:

$$loss_s = \sum_{i=1}^{N_{steps}} NLL(s_i, \hat{s}_i) \qquad (8)$$

Where NLL is shorthand for negative loss-likelihood and $N_{steps}$ is the length of the sequence. $Loss_p$ is the MSE of the predicted offsets w.r.t. the actual offsets. Following (Arnold and Yamazaki, 2019a), training computes connection weight gradients for both losses, and combines the losses by summing their signs at the connection level. This allows for combination of losses without adding hyperparameters for weight balancing. Weights are updated using the signSGD update rule (Bernstein et al., 2018), so gradient magnitudes are discarded. The learning rate is initialised to $5\times10^{-5}$. Every 10000 iterations, we measure prediction accuracy on the validation set, and reduce the learning rate by a factor 2 when the validation score has not improved in 50000 iterations. Training runs for 1.6M iterations, and we use the nets with the best prediction accuracy on the validation set for further evaluation and planning.

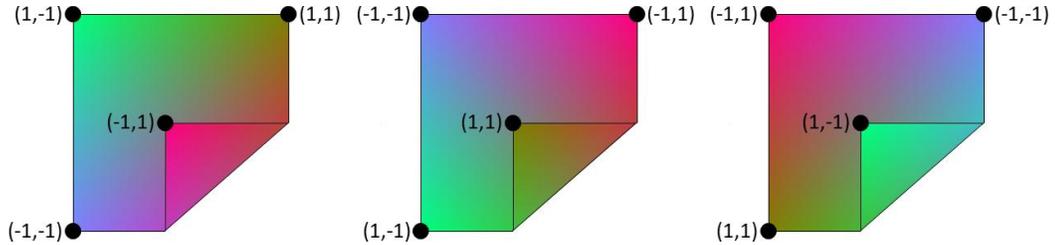

Figure 7. Equivalent mesh representations of a cloth shape. Geodesic coordinates are shown for the cloth corners and a gradient texture is added to visualise geodesic coordinates over the cloth surface. The representations differ in how they map geodesic space to Cartesian space, but represent the same shape. For every possible shape configuration of a square cloth, there are eight such equivalent mesh representations.

Mesh input is presented in a normalised form where the cloth in its fully spread, axis-aligned state runs from (-0.7, -0.7) to (0.7, 0.7) in the XY plane (this scale results from the way we set the voxel viewport, which is discussed in Section 4.7). Training employs various types of data augmentation. (1) Rotation and mirroring of manipulations sequences. (2) Introduction of non-manipulation examples. An example is converted into a non-manipulation example by changing the action into a null-manipulation, and changing the post-manipulation state into a copy of the pre-manipulation state. This is to learn the non-effect of null-manipulations. Recall that null-manipulations are represented as any action with a displacement vector shorter than 0.35 times the length of the side of the cloth. (3) Grasp point swaps. Grasp point order is immaterial, so swapping the grasp points results in a different representation of the exact same manipulation. Performing swaps ensures that the network learns the same outcome for both of these representations. (4) Addition of Gaussian noise with σ = 0.025 to vertex coordinates. (5) Conversion of input and target DMRs to any of their "Cartesian equivalent" representations. Figure 7 illustrates Cartesian equivalence. Each vertex in the mesh is identified by its geodesic $(u, v)$ coordinates. Recall that we let the geodesic coordinates run from [-1, -1] to [1, 1]. A shape configuration is given by a set of Cartesian coordinates $(x, y, z)$ for all vertices. However, for any shape configuration, there are multiple equivalent ways of assigning the Cartesian coordinates to the vertices. For example, when the cloth is laid out flat on the work surface, we may assign the top left corner to the vertex identified by geodesic coordinates (-1, -1), (-1, 1), (1, -1), or (1, 1). Each choice leaves us with two topologically coherent choices for the top right corner: if the top left corner is assigned to (-1, -1), the top right corner must be assigned to (1, -1) or (-1, 1). From there, all other assignments follow by necessity. Consequently, we have eight equivalent assignments of Cartesian coordinates to geodesic coordinates for each possible shape configuration. In data augmentation we convert states into randomly selected equivalent assignments, and convert manipulation inputs accordingly (recall that grasp point inputs are presented as geodesic coordinates).

Module M expects two grasp points for its manipulation input, but approximately one third of our manipulations are single-handed and thus have only one grasp point. To input these manipulations into the network, we feed the same grasp point on both inputs.

## 4.7. VOXEL-TO-MESH CONVERSION (VtM)

The second NN architecture is a network for converting voxel representations of cloth states to mesh representations (Arnold and Yamazaki, 2019b). We refer to this network as the Voxel-to-Mesh (VtM) net. This is a simple fully connected 7-layer network with input and output layer sizes matched to the voxel representation size and PMR size, and hidden layer size 4096 throughout. Input to the VtM net is given as a voxel representation of a cloth shape (centred in the voxel space). We use a normalised coordinate system in which the voxel viewport runs from (-1, -1, 0) to (1, 1, 0.125). As for pEM*D input, the fully spread, axis-aligned cloth runs from (-0.7, -0.7) to (0.7, 0.7) in this normalised viewport. This scaling is chosen so that any (intact) cloth shape fits within the viewport.

The voxelisation resolution is four times finer on the z-axis than on the x and y axes, to account for the fact that stable cloth shapes typically present detail such as wrinkles and layering on the z-axis, while spatial extension on the z-axis is limited compared to the x- and y-axes.

We denote the VtM net's functionality as follows:

$$VtM(s_i^v) = \dot{s}_i^p \tag{9}$$

Where $s_i^v$ is the voxel representation and $\dot{s}_i^p$ a probabilistic mesh estimate of the state at manipulation step $i$.

VtM produces PMRs, but for the planning procedure, we need a DMR, as deterministic graspable points can only be computed for DMRs. After estimating $\dot{s}_i^p$, we search for DMR $\dot{s}_i^d$ that is most (or at least highly) plausible w.r.t. $\dot{s}_i^p$. We refer to this procedure as *refinement* and denote it as follows:

$$R(\dot{s}_i^p, \tilde{s}_i^d) = \dot{s}_i^d \tag{10}$$

Where $\tilde{s}_i^d$ is a deterministic mesh used to initialise the search process. We can use the $\mu$ component of a previously obtained prediction for the current state when available, or use the $\mu$ component of $\dot{s}_i^p$. The DMR $\dot{s}_i^d$ is initialised to $\tilde{s}^d$, and iteratively refined to minimise the following loss measures:

- loss$_{nll}$: Negative log-likelihood of $\dot{s}_i^d$ w.r.t. $\dot{s}_i^p$.
- loss$_{spring}$: Spring energy.
- loss$_{up}$: upward bias loss.

To compute loss$_{spring}$, we define a set of springs between the vertices of the mesh, following a spring pattern common to cloth simulation (see e.g. (Choi & Ko, 2002)). Let $k$ be the distance between orthogonally neighbouring vertices. A vertex at indices $(u, v)$ in the cloth connects to neighbour vertices $(u, v \pm k)$ and $(u \pm k, v)$ ("stretch" springs), $(u \pm k, v \pm k)$ ("shear" springs), and $(u, v \pm 2k)$ and $(u \pm 2k, v)$ ("bend" springs), insofar these vertices exist. Spring energy loss is then calculated as follows.

$$loss_{spring} = \sum_{i=0}^{N_{springs}} (l_i - r_i)^2 \tag{11}$$

Here $l_i$ is the current length of the spring, $r_i$ is the resting length of the spring (i.e. its length in the cloth's fully spread-out default state), and $N_{springs}$ is the total number of springs.

Loss$_{up}$ serves merely to bias the optimisation against adjusting vertex positions downward, because such adjustments can push vertices into the working surface. The upward bias ensures that wrinkles produced by optimisation form in upward direction. This bias loss is computed as the mean over $max(0, \dot{s}_i^p.\mu_z - \dot{s}_i^d.z)$, where $\dot{s}_i^p.\mu_z$ and $\dot{s}_i^d.z$ are the $\mu_z$ and $z$ components of $\dot{s}_i^p$ and $\dot{s}_i^d$ and subtraction and *max* operate element-wise. Spring loss and upward bias loss are multiplied by 5000 and 1000, respectively, to be on the same order of magnitude as loss$_{nll}$. We update vertex positions using SignSGD on the compound loss, with an update step of 0.001. Optimisation runs until the loss stabilises or 300 iterations have passed.

Details such as small creases are hard to estimate from voxel representations. The PMR produced by VtM tends to miss such detail in its $\mu$ component. Where small creases or other details have disappeared, spring lengths computed over the $\mu$ component come out short, and $\sigma$ values are increased. Higher $\sigma$ values allow the optimisation step to move vertices further from their $\mu$ positions with smaller impact on loss$_{nll}$. Consequently, the optimisation step tends to fill in missing detail by moving vertices around so as to recover correct spring lengths, within the leeway provided by the locally increased $\sigma$ values. The resulting DMRs are thus more realistic cloth shapes than would be obtained by simply taking the $\mu$ component of the PMR. This realism is important for subsequent processing with the pEM*D net, as this net is trained on deterministic mesh data from the simulation. Skipping refinement was observed to significantly harm planning performance. However, refinement does not necessarily improve accuracy w.r.t. the ground truth compared to the $\mu$ component of the VtM estimate. A missing crease will often yield a smaller error than a crease recovered at a slight offset.

### 4.8. VtM Training

The VtM net is trained on the shape data from our simulation dataset. Meshes from the dataset are converted to voxel representation for input, by setting all voxels that contain at least one vertex to 1 and all other voxels to 0. We apply data augmentation by randomly rotating and mirroring meshes around the Z axis. Rotation in particular proved crucial to avoid excessive overfitting. Before conversion to voxel representation, we double the mesh resolution by interpolation. We then duplicate the resulting set of vertices and apply Gaussian noise to the vertex coordinates with $\sigma = 0.025$ in order to promote generalisation. This noise is absent in the ground truth, hence the net also learns to denoise its input. Finally we apply artificial self-occlusion by setting to 1 all voxel for which an occupied voxel with the same $x$ and $y$ index and a larger $z$ index exists. This occlusion approximates the self-occlusion that occurs for a single top-down camera viewpoint.

In Section 4.6 we discussed Cartesian equivalence between mesh representations. For effective VtM net training, it is crucial to account for this equivalence. As each shape configuration has eight equivalent mesh representations, there exist eight correct answers for each valid input to the VtM net. We found no principled way of designating any one answer as canonical. Simply using the meshes as-is as training targets leads to failure to train, because rotational augmentation, mirror augmentation, and repeated manipulation render the original assignment of Cartesian coordinates to geodesic coordinates completely opaque. Hence instead of enforcing a specific assignment, we allow all equivalent answers. We let the VtM net pick whichever assignment it finds convenient by training with the following loss:

$$loss_{VtM} = MIN\{NLL(s^j, \dot{s}_i^p) | j \in [0, ..., 7]\} \qquad (12)$$

Here $s^j$ is the $j^{th}$ Cartesian equivalent assignment in an arbitrary ordering of the ground truth's equivalent assignments. Training again uses the signSGD update rule with automated learning rate reduction as in pEM*D training, and is terminated when the learning rate drops below $10^{-7}$.

## 5. PLANNING LOGIC

Next we explain the planning algorithm. We perform planning through back-propagation, and address the problem of incomplete domain knowledge by introducing an epistemic uncertainty term in the loss function minimised by the planning procedure. Below we first present the planning algorithm, and then the loss function.

### 5.1. PLANNING ALGORITHM

Manipulation planning starts from a voxel representation of the current cloth state. When planning on basis of simulation data, we convert the DMR provided by the simulator to voxel format to emulate the real-world case. The voxel representation $s_0^v$ is converted to PMR $\dot{s}_0^p$ using the VtM net. Then we derive DMR $\dot{s}_0^d$ from $\dot{s}_0^p$ with the refinement routine described in Section 4G, using the $\mu$ component of $\dot{s}_0^p$ as initialisation value. We apply the grasp point detection routine on this DMR to obtain the set of graspable points for the current shape. On basis of $\dot{s}_0^d$ and DMR $s^*$ representing the goal state, we generate a manipulation plan. We assume the number of steps $n$ to be given and $\leq 3$. Generation of a plan of length $n$ is then performed as follows:

1. Initialise a random plan $m = <m_i, ..., m_{i+n-1}>$
2. Compute the planning loss (detailed below) for $m$.
3. If $n_{iterations}$ iterations have passed, return the plan with the lowest planning loss seen so far.
4. If the loss score has not improved for 5 steps, return to step 1.
5. Obtain loss gradients w.r.t. the manipulation inputs by means of backpropagation of the planning loss.
6. Adjust $m$ in accordance with these gradients using the iRprop- update rule (Igel and Hüsken, 2000), and return to step 2.

The number of search iteration is set in consideration of the plan length: $n_{iterations} = 25 + 25n$. Hyperparameters for iRprop- were set as follows: $\eta^+ = 1.2$, $\eta^- = 0.5$, $\Delta_{min} = 10^{-3}$, $\Delta_{max} = 0.05$. Update rates are initialised to 0.05. We run 256 planning instances in parallel, and adopt the plan with the lowest residual plan loss seen over the course of the planning processes in all instances.

The planning logic must account for the unusual action format. Recall that we have a variably-sized, discrete set of real-valued grasp point candidates for the current state, and that points not in this set should be considered ungraspable. In step (1), grasp points for the first manipulation in the plan are picked at random from the set of grasp points detected for the current state. These grasp points are not updated in step (6). Candidate grasp points for the first manipulation are known, so there is no need to search the input space aside from this discrete set. Search strains initialised with a bad choice of first-step grasp points usually fail to improve much, and quickly get reinitialised through step (4). For manipulations beyond the first, we only have (continually changing) probabilistic state predictions. This makes it hard to derive candidate grasp points, so we leave it up to the search process to figure out where viable grasp points are likely to appear. Note that non-first manipulations are never executed, so later grasp point positions need not be exact.

Due to prediction being less than perfect, different equivalent mesh representations of the same cloth shape may elicit slightly different plans. We compute all eight Cartesian equivalents of the current ($\dot{s}_i^d$) and goal DMR ($s^*$), and assign one eighth of the planning instances to each symmetry. For each non-first planning process in a trial, we initialise one instance per Cartesian equivalent with a modified version of the previously generated plan. We drop the first manipulation (as it has just been performed), snap the grasp points of the new first manipulation to the nearest points in the new set of candidate grasp points, and translate the resulting plan to each Cartesian equivalent.

Once a plan is generated, we check whether the system in fact expects the plan to produce a better approximation of the goal than the (estimated) current state provides, by comparing the error between the (estimated) current state and goal state with the error between the plan's predicted outcome and the goal state. If the latter exceeds the former, we set the next manipulation to a null-manipulation. The reason for this check is as follows. Sometimes the system achieves a good approximation before the final manipulation. When the system cannot find a plan that it expects to further improve upon the present state, it should ideally plan null-manipulations, but we find that it sometimes produces plans for which the expected outcome is slightly worse than the present state. By cancelling such manipulations, we avoid their potentially deleterious effects.

The generated plan is executed in simulation or hardware. After execution, we observe the resulting cloth state as a voxel representation again, and store the outcome. If the plan length was 1, we terminate the interaction. Otherwise, we repeat the planning process with plan length $n$ reduced by 1.

For subsequent steps of a multi-step manipulation process, we carry over some information from the preceding step. $EM^1D(\dot{s}_i^d, m_i)$ gives a probabilistic expectation $\hat{s}_{i+1}$ of the outcome of the first manipulation in the generated plan. For the subsequent plan generation process, we let $\dot{s}_{i+1}^d = R(VtM(s_{i+1}^v), \hat{s}_{i+1}.\mu)$. By using the $\mu$ component of the prediction for the new state as initial mesh for refinement, we aim to disambiguate states that are ambiguous in voxel representation. For example, if the cloth is folded neatly in two, determining which side of the resulting rectangle corresponds to the fold can be hard. This ambiguity would be expressed as high $\sigma$ values in $\dot{s}_{i+1}^p$ to allow for both interpretations. Refining from $\dot{s}_{i+1}^p.\mu$ could converge onto either interpretation. However, prediction $\hat{s}_{i+1}$ will place the fold unambiguously on one side or the other (whichever is consistent with the manipulation that produced $s_i$). Refining from $\hat{s}_{i+1}.\mu$ instead of $\dot{s}_{i+1}^p.\mu$ thus strongly biasing refinement towards the interpretation that is consistent with the state we predicted given the preceding manipulation. Prediction $\hat{s}_{i+1}$ also has a $\sigma$ component, but since the purpose of using the prediction as initialisation value is to disambiguate qualitatively different states, we have not sought to incorporate this $\sigma$ component into shape estimation.

A second way in which we carry through information from the preceding step is through memory trace $t_i$. At each subsequent plan generation process in a trial, we use memory trace $t_{i+1}$ obtained through $M(E(\dot{s}_i^d), t_i, m_i')$ as the memory trace input for the first instance of $M$. This similarly serves to allow preceding manipulations to influence plan generation. Given perfect knowledge of the current state, the current state and manipulation suffice to predict the next state. However, our state knowledge is not perfect. Knowledge about preceding states and manipulations can compensate for this imperfection to some extent (to consider an extreme case, given perfect knowledge of the preceding state and manipulation, we can compensate for a complete lack of knowledge of the current state through prediction). We expect that the memory trace captures some of this knowledge, in latent form, so we pass it on through subsequent runs of the planning process within a trial.

Computing $t_{i+1}$ requires some care given that, here too, we have to account for Cartesian equivalence. Recall that we seed plan search with a mixture of all equivalent representations. We translate the first manipulation of the generated plan for each symmetry and perform one forward propagation through pEM*D to obtain $t_{i+1}$ for all symmetries, and use corresponding pairs of input states and $t_{i+1}$ vectors for the next plan generation process.

5.2. PLANNING LOSS AND EPISTEMIC UNCERTAINTY AVOIDANCE

We address the problem of planning with incomplete domain knowledge using a dual network setup. Our approach relies on two mechanisms for uncertainty avoidance, one implicit and one explicit. Implicit avoidance is obtained by simply combining the prediction losses from two independently trained nets into a single planning loss. We can expect there to be fewer plans that deceive both nets into predicting an outcome close to the goal than plans that deceive just one net into doing so. Therefore, few plans leading through the unknown regions of the state-action space would produce a low combined prediction loss. Conversely, plans that stay within the known regions of the state-action space can be expected to elicit similar predictions from both nets, so a valid plan for obtaining the goal state should elicit a low combined prediction loss. We include this implicit uncertainty avoidance strategy as a baseline in our experiments.

This implicit strategy relies on the fact that discrepancy between predictions will raise the combined prediction loss. However, various other factors affect the combined prediction loss, making it a poor indicator of epistemic uncertainty as such. Given two nets, we can also calculate an explicit measure of prediction discrepancy, which is more specific and informative about the epistemic uncertainty of a given plan. In particular, we can let the explicit measure take into account prediction discrepancy between non-final cloth states predicted to occur between subsequent manipulations. Plans for which both nets predict similar final outcomes will still be unreliable if predictions for these intermediate states diverge.

We introduce an epistemic uncertainty term in the loss function. Planning loss for dual-net planning with $net_a$ net and $net_b$ is defined as follows:

$$loss_{dual} = \frac{NLL(w(s^*), w(\hat{s}_n^a)) - low_n^a}{high_n^a - low_n^a} + \frac{NLL(w(s^*), w(\hat{s}_n^b)) - low_n^b}{high_n^b - low_n^b} + \alpha \cdot u \tag{13}$$

Here $s^*$ is the goal state, $\hat{s}_n^j$ is $net_j$'s predicted outcome for the current plan, $high_n^j$ and $low_n^j$ are reference values for normalising the losses from each net, $\alpha$ is a fixed weight, and $u$ is the epistemic uncertainty for the current plan (explained in detail below). We refer to the first two terms as accuracy terms and to the last term as the uncertainty term. Function $w$ is a simple whitening operation, rescaling its input state such that its vertices have a mean of 0 and a standard deviation of 1 for each Cartesian dimension. For PMRs this is over the $\mu$ elements, with $\sigma$ values rescaled accordingly. This serves to eliminate the effect of the size of a cloth shape on the loss. A tightly crumpled shape has small coordinate values overall. Without whitening, such shapes produce relatively small losses against topologically dissimilar states compared to more spatially extended shapes. Whitening makes size irrelevant and focuses loss on shape similarity, which appeared to slightly improve results in preliminary experiments. Figure 8 illustrates the information flow in the dual-net compound architecture.

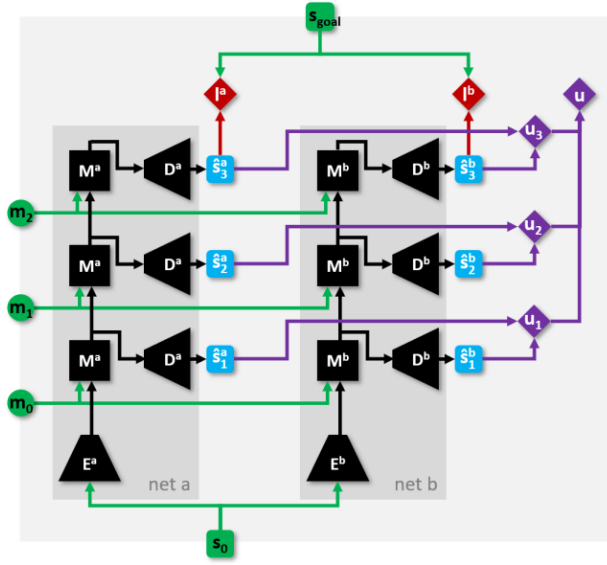

Figure 8. Dual network architecture with explicit epistemic uncertainty quantification, rolled out for 3-step plans. Green items indicate network inputs and (externally supplied) goal state. Black items are neural network modules. Blue items are state predictions. Red items are losses. Purple items are epistemic uncertainty values.

Normalising the individual terms of the planning loss is not strictly necessary, but makes it easier to balance the accuracy terms with the uncertainty term. It also lets us avoid the need to retune $\alpha$ when changing out the networks, which allows for fairer comparison experiments (we will be comparing performance against a purely voxel-based variant below).

Note that normalisation is only meaningful when combining multiple loss terms, as our planning algorithm employs iRprop-, which uses the sign, not the magnitude, of the loss gradients. Also, subtraction of $low_n^j$ in the numerators of the loss terms in Equation 13 does not affect the search process, but makes search loss values easier to interpret.

The reference values $high_n^j$ and $low_n^j$ for accuracy term normalisation are computed as follows.

$$low_n^j = Median\left\{NLL\left[w(s_{i+n}^d), w\left(EM^nD(s_i^d, m_{i:i+n-1}^d)\right)\right] \middle| d \in [0 \dots N_v] \wedge i \in [0 \dots n-3]\right\} \quad (14)$$

$$high_n^j = Median\left\{NLL\left[w(s_{i+n}^d), w\left(EM^nD(s_i^d, R^m(n))\right)\right] \middle| d \in [0 \dots N_v] \wedge i \in [0 \dots n-3]\right\}$$

Here $N_v$ is the number of examples in the validation set (100 here), $d$ indexes the example sequences in the dataset, $i$ indexes the steps within an example sequence, $R^m(n)$ generates random manipulation sequences of length $n$ (disregarding the restrictions used in data generation), and *Median* computes the median over its argument set. As for the range of $i$, recall that our example manipulation sequences are of length 3 (resulting in state sequences of length 4). We source example sequences of length $n$ by selecting all subsequences indexed by $i: i + n - 1$ (inclusive) with $i \in [0, 3 - n]$. Hence for length $n$ we find $4 - n$ sequences per example.

These values give the range of loss values we can expect to see during successful plan search. Value $high_n^j$ gives the typical loss value for randomly initialised plans, while $low_n^j$ gives the typical residual loss we can expect to see for plans that in fact produce the goal state.

Epistemic uncertainty value $u$ serves to let the search process avoid unfamiliar regions of the state-action space. We explain its reasoning and calculation below. The pEM*D net provides predictions for the plans considered during plan search, including aleatoric uncertainty for the outcome and (if we decode them) intermediate states. However, these predictions will only be reliable for plans that fall within the regions of the state-action space that were sampled by the dataset. Since our dataset samples only part of the state-action space, prediction reliability will vary across plans. Uncertainty about prediction reliability is not captured by the (aleatoric) uncertainty estimates in the predictions, so we assess epistemic uncertainty separately. Epistemic uncertainty value $u$ is calculated as follows:

$$u = \sum_{i=1}^{n} \frac{e^{\beta(u_i - \gamma)}}{n} \qquad u_i = \frac{diff(\hat{s}_i^a, \hat{s}_i^b) - \mu_i^v}{\sigma_i^v} \quad (15)$$

State $\hat{s}_i^j$ is *net$_j$*'s prediction of the cloth state at step $i$ of the current plan, obtained by decoding the latent state $\hat{c}_i^j$ obtained when the current plan and state are fed into *net$_j$*. Values $\mu_i^v$ and $\sigma_i^v$ are the mean and standard deviation of the difference between the two nets' predictions at the $i^{th}$ state of their prediction sequences, computed over the validation set. This gives an estimate of the distribution of the difference between predictions for unseen examples drawn from the region of the state-action space covered by the training data. Parameters $\beta$ and $\gamma$ tune how uncertainty impacts on the planning process. We set $\beta$ to 5 and $\gamma$ to 1.25 in our experiments. With these settings, planning is strongly averse to prediction discrepancies exceeding the expected discrepancy by over 1.25 standard deviations. The *diff* function quantifies the difference between two predictions. This requires some consideration, because whether or not a given network output can reasonably be interpreted as a prediction depends on whether the network inputs lie within (or near enough to) the known region of the state-action space. For inputs outside the known regions of the state-action space, we should regard network output as nonsense. For this reason, we let the *diff* function compare its arguments simply as arrays of real values, instead of interpretating them as PMRs. We let *diff* whiten its arguments and compute their MSE.

Implicit and explicit uncertainty avoidance can be expected to affect plan search in subtly different ways. Consider a situation where we have two possible plan updates at a given step in the plan generation process. Update *x* improves prediction loss for net *a*, leaves prediction loss for net *b* roughly unchanged, and increases the difference between the predictions. Update *y* leaves prediction loss for both nets roughly unchanged, but brings their predictions closer together. The implicit strategy would always prefer update *x*, while in cases where the discrepancy is concerningly high (as tuned with the loss balancing parameters), the explicit strategy would prefer update *y*. This suggests that the implicit strategy would be less effective at preventing the planning process from wandering into unknown territory. The implicit strategy may suffice to save the system from being thoroughly deceived by deceptive plans, but if it spends more update cycles in unknown regions (where gradients are unreliable), then we should still expect it to produce worse outcomes than the explicit strategy.

Our experiments include single-net planning as a baseline. Single-net planning does not account for epistemic uncertainty (neither implicitly nor explicitly). The planning loss for single-net planning is $NLL\big(w(s^*), w(\hat{s}_n)\big)$, with $\hat{s}_n$ the predicted plan outcome.

## 6. EXPERIMENTS

We present three categories of results: estimation accuracy, prediction accuracy, and planning accuracy.

### 6.1. ESTIMATION

First we evaluate the VtM net's shape estimation accuracy. This evaluation omits the first state ($s_0$) of each data sequence, as this state is always the flat default state and including it would artificially improve scores. Estimation accuracy is shown in Table 2. Scores indicate Euclidean distance between vertices' estimated and actual positions, with the length of the cloth as unit. For the initial estimates (PMRs), scores are computed using the $\mu$ component. As noted above, refinement is intended to recover realism, and does not necessarily improve quantitative accuracy compared to the $\mu$ component of the initial estimate. The scores reveal some overfitting, and some errors in cloth layer ordering on the z-axis are observed, as well as instances of self-intersection in more complex cloth shapes, which is unsurprising as the refinement routine does not perform collision checking. Self-intersection would be problematic for further processing in simulation, but not for processing by the pEM*D net (note that self-intersection also occurs in the shapes the pEM*D net observes during training, due to noise applied in data augmentation). Examples of VtM estimates and their refinements are shown in Figure 9. We observe good approximation of global shape in the initial estimate, and recovery or fill-in of fine detail by refinement. Average time cost of mesh estimation is 3.7 milliseconds for the forward VtM pass generating the initial estimation and 0.57 seconds for refinement.

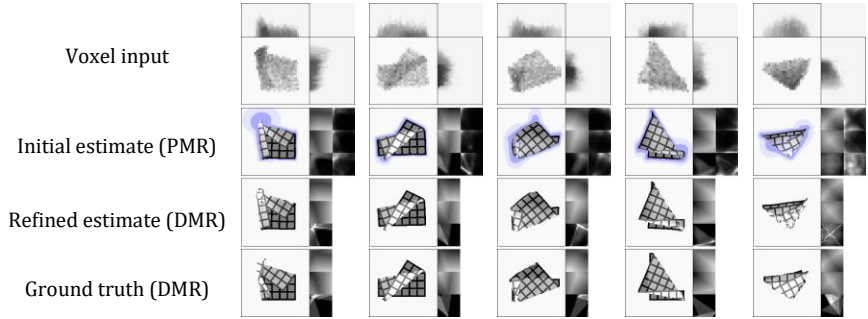

Figure 9. Representative examples of shape estimation and refinement (test set data). Each column represents one example. The last example shows a case where the z-ordering of the cloth layers is particularly difficult to infer from the voxel representation, leading to ambiguous z-ordering in the estimate.

TABLE 2. MESH ESTIMATION ACCURACY

| Unit: Length of the cloth | | | Step | | | |
|---|---|---|---|---|---|---|
| | | | 1 | 2 | 3 | all |
| Test (100 seq.) | PMR (μ-component) | mean | 0.0142 | 0.0481 | 0.0800 | 0.0474 |
| | | (st.dev.) | (0.0201) | (0.0315) | (0.0383) | (0.0410) |
| | | median | 0.0116 | 0.0417 | 0.0781 | 0.0357 |
| | DMR (refined) | mean | 0.0137 | 0.0488 | 0.0833 | 0.0486 |
| | | (st.dev.) | (0.0198) | (0.0340) | (0.0410) | (0.0434) |
| | | median | 0.0114 | 0.0408 | 0.0831 | 0.0349 |
| Train (100 seq.) | PMR (μ-component) | mean | 0.0143 | 0.0418 | 0.0662 | 0.0408 |
| | | (st.dev.) | (0.00846) | (0.0242) | (0.0289) | (0.0308) |
| | | median | 0.0124 | 0.0354 | 0.0625 | 0.0331 |
| | DMR (refined) | mean | 0.0135 | 0.0420 | 0.0680 | 0.0411 |
| | | (st.dev.) | (0.00674) | (0.0255) | (0.0307) | (0.0323) |
| | | median | 0.0122 | 0.0342 | 0.0617 | 0.0315 |

To assess the quality of the uncertainty estimations (σ component of the estimate), we divide errors by the corresponding σ values. Under perfect uncertainty estimation, this value would be $\sqrt{2/\pi} \approx 0.798$ (the ratio between mean deviation and standard deviation for multivariate normal distributions). We obtain a median value of 0.770 for the test set, indicating that the σ component of the generated meshes represents the actual uncertainty well, overestimating it slightly.

6.2. PREDICTION

Next we evaluate the pEM*D net's prediction ability. For purpose of dual-net planning we independently trained two identical networks. For the evaluation here and for single-net planning in the next section we use the net with the best validation score. We include two baselines:

- Baseline A1 (No VtM): Prediction without shape estimation (i.e. prediction using the initial state's ground truth as input).

- Baseline A2 (Voxel-based): Prediction by an EM*D net trained to predict in voxel format (similar to previous work (Arnold and Yamazaki, 2019a)).

The voxel baseline requires some modification of the encoder and decoder modules, due to the different state input format. Following (Arnold and Yamazaki, 2019a), the encoder and decoder modules for the voxel-based EM*D nets are composed of 6 3D-convolutional layers each, with 32, 32, 64, 128, 256, and 512 feature maps (order reversed for decoder). Kernel size is 3×3×3 throughout. We use striding to reduce resolution at each layer, with a stride of 2 for all dimensions in all layers, except for the z-dimension on the first layer (last layer in the decoder), which has a stride of 1. We also tested a voxel-based EM*D with densely connected layers (as used in the mesh version), but found convolution layers to perform better in the voxel version. The M module is identical to the mesh version. Voxel representations are in the same format as used for VtM net input in the mesh-based system, with the same 32×32×16 resolution. As voxel representations are input to the planning network directly, the VtM net is not used in the voxel baseline.

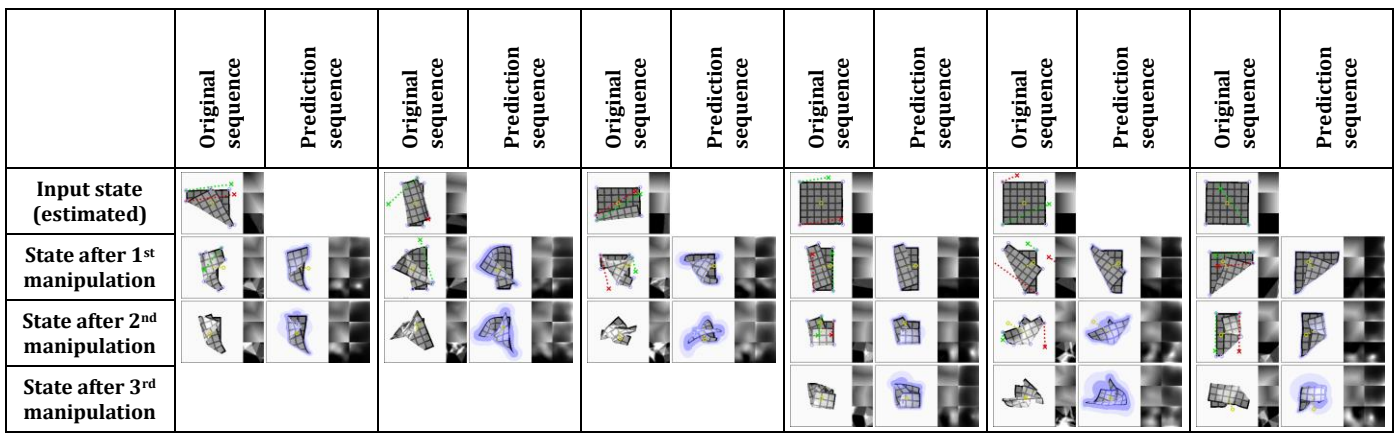

Figure 10. Example prediction results for sequences of two and three manipulations (main experiment, test set data). Input for prediction is an estimation of the original sequence's first state and the sequence of manipulation inputs. Red and green dotted lines show manipulation trajectories. Lilac circles mark grasp point candidates. Yellow dotted lines on predictions indicate the predicted displacement of the cloth's centre point. Trajectories that extend outside the viewport are wrapped around the border for visualisation purposes. The last example shows a case where the system fails to predict that the shape partially unfolds during the last manipulation.

TABLE 3. SHAPE PREDICTION ACCURACY (MESH FORMAT)

| Unit: Length of the cloth | | | Sequence length | | | | | | | |
|---|---|---|---|---|---|---|---|---|---|---|
| | | | 1 | | | | 2 | | | 3 |
| | | | Start point | | | | Start point | | | Start point |
| | | | 0 | 1 | 2 | all | 0 | 1 | all | 0 / all |
| Test data (100 seq.) | Main | mean | 0.0161 | 0.0451 | 0.0914 | 0.0509 | 0.0419 | 0.0837 | 0.0628 | 0.0809 |
| | | (st.dev.) | (0.00589) | (0.0244) | (0.0431) | (0.0423) | (0.0206) | (0.0364) | (0.0362) | (0.0347) |
| | | median | 0.01460 | 0.0381 | 0.0857 | 0.0362 | 0.0373 | 0.0753 | 0.0513 | 0.0737 |
| | Baseline A1: No VtM | mean | 0.0158 | 0.0412 | 0.0712 | 0.0428 | 0.0417 | 0.0811 | 0.0614 | 0.0810 |
| | | (st.dev.) | (0.00548) | (0.0214) | (0.0340) | (0.0325) | (0.0206) | (0.0355) | (0.0351) | (0.0347) |
| | | median | 0.0138 | 0.0349 | 0.0616 | 0.0325 | 0.0368 | 0.0719 | 0.0502 | 0.0738 |
| Training data (100 seq.) | Main | mean | 0.0163 | 0.0402 | 0.0773 | 0.0446 | 0.0389 | 0.0662 | 0.0525 | 0.0627 |
| | | (st.dev.) | (0.00726) | (0.0206) | (0.0391) | (0.0360) | (0.0188) | (0.0294) | (0.0282) | (0.0263) |
| | | median | 0.0137 | 0.0343 | 0.0673 | 0.0338 | 0.0332 | 0.0566 | 0.0453 | 0.0562 |
| | Baseline A1: No VtM | mean | 0.0162 | 0.0378 | 0.0571 | 0.0370 | 0.0386 | 0.0624 | 0.0505 | 0.0626 |
| | | (st.dev.) | (0.00728) | (0.0195) | (0.0274) | (0.0259) | (0.0186) | (0.0267) | (0.0259) | (0.0264) |
| | | median | 0.0136 | 0.0330 | 0.0491 | 0.0313 | 0.0332 | 0.0554 | 0.0442 | 0.0557 |

TABLE 4. PREDICTION ACCURACY (VOXEL FORMAT)

| Unit: mean voxel value difference | | | Sequence length | | | | | | | |
|---|---|---|---|---|---|---|---|---|---|---|
| | | | 1 | | | | 2 | | | 3 |
| | | | Start point | | | | Start point | | | Start point |
| | | | 0 | 1 | 2 | all | 0 | 1 | all | 0 / all |
| Test data (100 seq.) | Main | mean | 0.00858 | 0.0102 | 0.0131 | 0.0106 | 0.0102 | 0.0134 | 0.0118 | 0.0135 |
| | | (st.dev.) | (0.00284) | (0.00230) | (0.00289) | (0.00328) | (0.00232) | (0.00273) | (0.00301) | (0.00280) |
| | | median | 0.00833 | 0.00996 | 0.0126 | 0.0105 | 0.0101 | 0.0131 | 0.0114 | 0.0132 |
| | Baseline A2: Voxel-based | mean | 0.00893 | 0.02069 | 0.0316 | 0.0204 | 0.0200 | 0.0326 | 0.0263 | 0.0325 |
| | | (st.dev.) | (0.00267) | (0.00607) | (0.00936) | (0.0114) | (0.00598) | (0.00891) | (0.00986) | (0.00885) |
| | | median | 0.00866 | 0.0202 | 0.0296 | 0.0192 | 0.0193 | 0.0329 | 0.0242 | 0.0330 |
| Training data (100 seq.) | Main | mean | 0.00782 | 0.00959 | 0.0121 | 0.00983 | 0.00958 | 0.0121 | 0.0109 | 0.0120 |
| | | (st.dev.) | (0.00262) | (0.00223) | (0.00266) | (0.00306) | (0.00223) | (0.00261) | (0.00275) | (0.00258) |
| | | median | 0.00744 | 0.00934 | 0.0116 | 0.00962 | 0.00920 | 0.0116 | 0.0103 | 0.0115 |
| | Baseline A2: Voxel-based | mean | 0.00911 | 0.0195 | 0.0272 | 0.0186 | 0.0187 | 0.0267 | 0.0227 | 0.0266 |
| | | (st.dev.) | (0.00279) | (0.00572) | (0.00607) | (0.00899) | (0.00537) | (0.00560) | (0.00678) | (0.00544) |
| | | median | 0.00866 | 0.0189 | 0.0270 | 0.0180 | 0.0177 | 0.0259 | 0.0220 | 0.0263 |

TABLE 5. POSITION PREDICTION ACCURACY

| Unit: Length of the cloth | | | Sequence length | | |
|---|---|---|---|---|---|
| | | | 1 | 2 | 3 |
| Test data (100 seq.) | Main | mean | 0.0295 | 0.0383 | 0.0505 |
| | | (st.dev.) | (0.0352) | (0.0346) | (0.0407) |
| | | median | 0.0174 | 0.0282 | 0.0353 |
| | Baseline A1: No VtM | mean | 0.0269 | 0.0379 | 0.0506 |
| | | (st.dev.) | (0.0295) | (0.0349) | (0.0403) |
| | | median | 0.0163 | 0.0259 | 0.0370 |
| Training data (100 seq.) | Main | mean | 0.0235 | 0.0214 | 0.0205 |
| | | (st.dev.) | (0.0262) | (0.0175) | (0.0115) |
| | | median | 0.0147 | 0.0189 | 0.0192 |
| | Baseline A1: No VtM | mean | 0.0164 | 0.0197 | 0.0201 |
| | | (st.dev.) | (0.0114) | (0.0108) | (0.0114) |
| | | median | 0.0138 | 0.0180 | 0.0180 |

Accuracies for prediction of the final outcome of manipulation sequences of length one, two, and three are given in Tables 3 (mesh format) and 4 (voxel format). Evaluation is over the full test set (100 sequences) and 100 sequences from the training set. As is to be expected, the prediction error increases with the length of the manipulation sequence. Comparing the scores for training and test data, some overfitting is again apparent. Score differences between the main experiment and baseline A1 (no VtM) are small, indicating that shape estimation is mostly sufficient to bring out the system's predictive potential. Example predictions are shown in Figure 10. We observe that precision drops over the course of the sequence, but the global cloth shape remains clearly recognisable for 3-step prediction. Increasing uncertainty over steps is expressed in increasing σ values over the course of the sequence (lilac shading on predicted shapes).

Again we evaluate uncertainty estimates by dividing the error for each predicted vertex coordinate by the corresponding sigma value, and averaging the results. We obtain median values of 0.750, 0.785 and 0.861 for 1-, 2-, and 3-step prediction, respectively, indicating decent uncertainty estimation (recall that the ideal value is 0.798).

We also measure accuracy of the predicted cloth centre position shift for the main system configuration, and when operating with ground truth meshes as input (baseline A1). Results are given in Table 5. We see that position shift can be predicted with decent accuracy, which means we can calculate the approximate position of the predicted cloth shapes in the workspace. We make no use of this here, but this functionality would be important in many practical scenarios with workspace and arm reach limitations.

To evaluate baseline A2 (voxel-based), we need to compare shape prediction ability across shape representation formats. For this purpose, we convert predictions from the mesh-based main experiment into voxel format as follows. For each vertex $i$, we draw 1024 random samples from the multivariate normal distribution given for vertex $i$ by the prediction. We bin the samples into voxels, and then divide the voxel values by 1024. This results in a voxel-format prediction for the single vertex, consisting of values $p^i_{xyz}$ quantifying the probability of vertex $i$ falling in the voxel with indices $(x, y, z)$. We combine the single-vertex predictions into a full-cloth prediction by calculating for each voxel the probability that one or more vertices fall into it as follows:

$$p_{xyz} = 1 - \prod_{i=0}^{n_{vertices}} \left(1 - p^i_{xyz}\right) \quad (16)$$

We now have occupancy probabilities for all voxels, which together constitute a cloth-shape prediction in voxel format. This format corresponds to the state format of the voxel version of the planning system, allowing for direct accuracy comparison. Results are given in Table 4. The error unit here is the mean voxel value difference. The ground truth is a binary voxel representation (1 for voxels occupied by the cloth, 0 for empty voxels). We observe that mesh-based prediction outperforms voxel-based prediction by a good margin, despite the fact that the mesh-based predictions pass through two noisy format conversions in this evaluation (voxel to mesh conversion on the input states, and mesh to voxel conversion on the output states). State input to both systems is the same, but recall that the mesh-based system takes grasp point input in geodesic coordinates. As there exists no geodesic space in the voxel-based system, it takes grasp points in Cartesian coordinates.

6.3. PLANNING (SIMULATION)

Next we evaluate planning performance. In addition to the main system, we again evaluate a number of baselines in order to assess the contributions of specific aspects.

- Baseline B1: No VtM. In this baseline we perform dual-net planning on basis of ground truth meshes instead of the VtM net's estimates thereof. This provides an upper bound to how well the planning system would perform with perfect knowledge of the current cloth shape, which allows us to assess how well the VtM net performs in the context of the full planning system.

- Baseline B2: Single-net planning. This baseline lets us isolate the contribution of our dual-net epistemic uncertainty avoidance strategy.

- Baseline B3: Dual-net planning without explicit epistemic uncertainty loss. The use of two nets instead of one can be expected produce some level of epistemic uncertainty avoidance, as discussed in Section 5.2. This baseline lets us isolate the contribution of explicit epistemic uncertainty measure $u$ in our uncertainty avoidance strategy. This baseline is obtained by setting $\alpha$ to 0 in Equation 13.

- Baseline B4: Voxel-based dual-net planning. For comparison with the voxel-based approach in (Arnold and Yamazaki, 2019a), we perform planning with EM*D nets operating on voxel representations directly. This baseline adds dual-net planning functionality to the voxel version, to facilitate evaluation of the effect of the representation format in comparison to the main experiment.

- Baseline B5: Voxel-based single-net planning. This baseline most closely approximates (Arnold and Yamazaki, 2019a).

We evaluate planning ability on all 100 examples from the test data set and 100 examples from the training data set. We use all sequences of length one to three that can be sourced from this set of examples. For 1-step sequences, sequences may start at the first ($s_0$), second ($s_1$), or third ($s_2$) state in the sequence. For 2-step sequence, we may start at the first or second state. Note that as our data sequences are of length three, 3-step planning always starts from the first (spread out) state. The same sequences are used for each experiment. Table 6 shows accuracy of the outcomes of interleaved planning and execution measured as the Euclidean distance between goal and outcome for each mesh vertex, with the length of the side of the cloth as unit. Our planning conditions aim to optimise oriented shape, while placing no constraints on the position of the obtained shape in space. Hence for evaluation, we first align the outcome and goal shapes in space. We do this by computing offset vectors for all vertices (i.e. the difference between a vertex's position in the goal shape and the outcome shape), and subtracting the average offset from all vertices, thereby obtaining the best alignment that can be achieved through translation. Scatter plots in Figure 11 show the scores for all examples from the test set for the main experiment and baseline B1, plotting planning accuracy against prediction accuracy as evaluated in Section 6.2. Figure 12 shows examples of planning and execution sessions. Note that many goal shapes can be produced through more than one manipulation sequence. In a number of the multi-step examples shown in Figure 12, we see that the system invented manipulation sequences that differ from the original sequence, yet closely approximate the same outcome. Parallel instances of the planning process, as well as repeated runs of the plan generation process, can produce different valid plans.

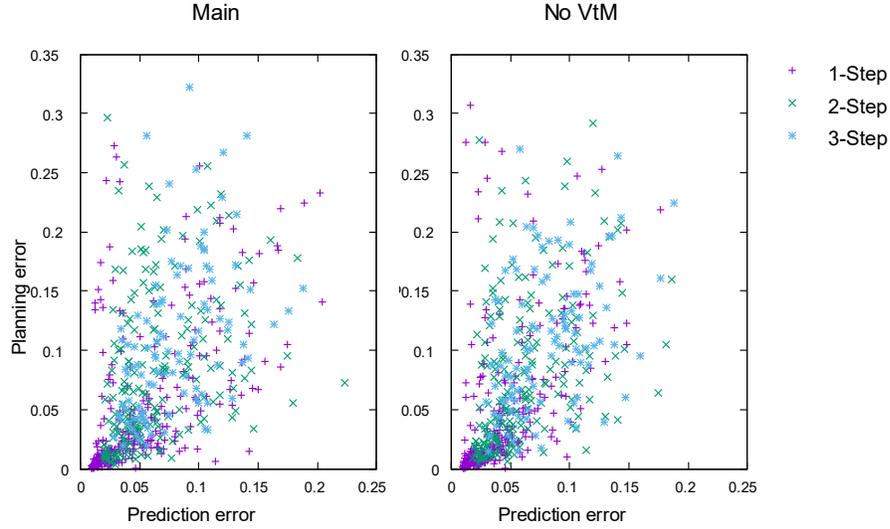

Figure 11. Planning accuracy plotted against prediction accuracy for all 1-, 2-, and 3-step sequences in the test set. Left: Main experiment. Right: Baseline B1 (No VtM). Error unit: length of the cloth. For predictions, errors are calculated for the $\mu$ component of the prediction. Unit is the length of the side of the cloth.

TABLE 6. PLANNING ACCURACY (MESH-BASED EVALUATION)

| Unit: Length of the cloth | | | Sequence length | | | | | | | |
|---|---|---|---|---|---|---|---|---|---|---|
| | | | 1 | | | | 2 | | | 3 |
| | | | Start point | | | | Start point | | | Start point |
| | | | 0 | 1 | 2 | All | 0 | 1 | All | 0 / all |
| Test data (100 seq.) | *Main* | mean | 0.0291 | 0.0411 | 0.0783 | 0.0495 | 0.0725 | 0.0913 | 0.0819 | 0.109 |
| | | (st.dev) | (0.0527) | (0.0499) | (0.0637) | (0.0596) | (0.0680) | (0.0596) | (0.0646) | (0.0652) |
| | | median | 0.00566 | 0.0223 | 0.0591 | 0.0232 | 0.0453 | 0.0819 | 0.0643 | 0.0969 |
| | *Baseline B1: no VtM* | mean | 0.0288 | 0.0395 | 0.0718 | 0.0467 | 0.0678 | 0.0905 | 0.0792 | 0.109 |
| | | (st.dev) | (0.0591) | (0.0501) | (0.0637) | (0.0607) | (0.0596) | (0.0639) | (0.0628) | (0.0575) |
| | | median | 0.00565 | 0.0193 | 0.0488 | 0.0185 | 0.0495 | 0.0803 | 0.0619 | 0.102 |
| | *Baseline B2: single-net* | mean | 0.0333 | 0.0402 | 0.0778 | 0.0504 | 0.0809 | 0.0986 | 0.0898 | 0.126 |
| | | (st.dev) | (0.0640) | (0.0425) | (0.0629) | (0.0606) | (0.0611) | (0.0595) | (0.0610) | (0.0712) |
| | | median | 0.00654 | 0.0259 | 0.0550 | 0.0259 | 0.0640 | 0.0917 | 0.0819 | 0.123 |
| | *Baseline B3: dual-net without uncertainty loss* | mean | 0.0314 | 0.0408 | 0.0677 | 0.0466 | 0.0813 | 0.0985 | 0.0899 | 0.121 |
| | | (st.dev) | (0.0577) | (0.0485) | (0.0510) | (0.0547) | (0.0736) | (0.0602) | (0.0678) | (0.0665) |
| | | median | 0.00550 | 0.0246 | 0.0588 | 0.0252 | 0.0598 | 0.0844 | 0.0726 | 0.124 |
| | *Baseline B4: Voxel / dual* | mean | 0.0278 | 0.0635 | 0.0915 | 0.0609 | 0.112 | 0.132 | 0.122 | 0.169 |
| | | (st.dev) | (0.0336) | (0.0531) | (0.0669) | (0.0590) | (0.0764) | (0.0663) | (0.0722) | (0.0599) |
| | | median | 0.0140 | 0.0481 | 0.0689 | 0.0370 | 0.0887 | 0.127 | 0.114 | 0.181 |
| | *Baseline B5: Voxel / single* | mean | 0.0319 | 0.0790 | 0.0942 | 0.0683 | 0.132 | 0.162 | 0.147 | 0.192 |
| | | (st.dev) | (0.0502) | (0.0668) | (0.0655) | (0.0668) | (0.0752) | (0.0712) | (0.0748) | (0.0590) |
| | | median | 0.0129 | 0.0546 | 0.0782 | 0.0445 | 0.127 | 0.168 | 0.156 | 0.205 |
| Training data (100 seq.) | *Main* | mean | 0.0384 | 0.0407 | 0.0718 | 0.0503 | 0.0714 | 0.0902 | 0.0808 | 0.119 |
| | | (st.dev) | (0.0605) | (0.0488) | (0.0640) | (0.0601) | (0.0529) | (0.0583) | (0.0565) | (0.0612) |
| | | median | 0.00625 | 0.0251 | 0.0423 | 0.0267 | 0.0481 | 0.0770 | 0.0636 | 0.115 |
| | *Baseline B1: no VtM* | mean | 0.0417 | 0.0343 | 0.0631 | 0.0464 | 0.0731 | 0.0867 | 0.0799 | 0.109 |
| | | (st.dev) | (0.0743) | (0.0397) | (0.0618) | (0.0615) | (0.0614) | (0.0659) | (0.0641) | (0.0570) |
| | | median | 0.00639 | 0.0186 | 0.0334 | 0.0203 | 0.0462 | 0.0657 | 0.0582 | 0.0992 |
| | *Baseline B2: single-net* | mean | 0.0426 | 0.0345 | 0.0687 | 0.0486 | 0.0925 | 0.0958 | 0.0942 | 0.127 |
| | | (st.dev) | (0.0654) | (0.0379) | (0.0572) | (0.0566) | (0.0692) | (0.0655) | (0.0674) | (0.0667) |
| | | median | 0.00709 | 0.0191 | 0.0491 | 0.0248 | 0.0867 | 0.0853 | 0.0862 | 0.118 |
| | *Baseline B3: dual-net without uncertainty loss* | mean | 0.0427 | 0.0372 | 0.0694 | 0.0495 | 0.0881 | 0.0834 | 0.0857 | 0.131 |
| | | (st.dev) | (0.0650) | (0.0449) | (0.0595) | (0.0588) | (0.0692) | (0.0532) | (0.0617) | (0.0682) |
| | | median | 0.00717 | 0.0212 | 0.0453 | 0.0251 | 0.0741 | 0.0725 | 0.0725 | 0.129 |
| | *Baseline B4: Voxel / dual* | mean | 0.0383 | 0.0622 | 0.0876 | 0.0627 | 0.113 | 0.132 | 0.123 | 0.169 |
| | | (st.dev) | (0.0446) | (0.0518) | (0.0673) | (0.0589) | (0.0679) | (0.0676) | (0.0684) | (0.0574) |
| | | median | 0.0166 | 0.0475 | 0.0652 | 0.0420 | 0.105 | 0.133 | 0.119 | 0.172 |
| | *Baseline B5: Voxel / single* | mean | 0.0514 | 0.0782 | 0.0835 | 0.0711 | 0.155 | 0.154 | 0.155 | 0.202 |
| | | (st.dev) | (0.0698) | (0.0741) | (0.0676) | (0.0719) | (0.0813) | (0.0714) | (0.0765) | (0.0543) |
| | | median | 0.0179 | 0.0503 | 0.0593 | 0.0435 | 0.159 | 0.158 | 0.159 | 0.207 |

TABLE 7. PLANNING ACCURACY (VOXEL-BASED EVALUATION)

| Unit: mean voxel value difference | | | Sequence length | | | | | | | |
|---|---|---|---|---|---|---|---|---|---|---|
| | | | 1 | | | | 2 | | | 3 |
| | | | Start point | | | | Start point | | | Start point |
| | | | 0 | 1 | 2 | all | 0 | 1 | all | 0 / all |
| Test data (100 seq.) | Main | mean | 0.00282 | 0.00516 | 0.00885 | 0.00561 | 0.00783 | 0.0103 | 0.00907 | 0.0119 |
| | | (st.dev) | (0.00347) | (0.00427) | (0.00501) | (0.00496) | (0.00511) | (0.00484) | (0.00512) | (0.00507) |
| | | median | 0.00150 | 0.00394 | 0.00793 | 0.00412 | 0.00662 | 0.00964 | 0.00836 | 0.0116 |
| | Baseline B4: Voxel / dual | mean | 0.00768 | 0.0182 | 0.0254 | 0.0171 | 0.0239 | 0.0308 | 0.0273 | 0.0336 |
| | | (st.dev) | (0.00722) | (0.00985) | (0.0117) | (0.0122) | (0.0114) | (0.0112) | (0.0118) | (0.00866) |
| | | median | 0.00504 | 0.0166 | 0.0243 | 0.0146 | 0.0222 | 0.0305 | 0.0264 | 0.0332 |
| | Baseline B5: Voxel / single | mean | 0.00818 | 0.0211 | 0.0269 | 0.0187 | 0.0282 | 0.0361 | 0.0322 | 0.0412 |
| | | (st.dev) | (0.00810) | (0.0117) | (0.0127) | (0.0135) | (0.0127) | (0.0134) | (0.0136) | (0.0122) |
| | | median | 0.00455 | 0.0203 | 0.0249 | 0.0170 | 0.0289 | 0.0380 | 0.0319 | 0.0420 |
| Training data (100 seq.) | Main | mean | 0.00323 | 0.00523 | 0.008302 | 0.00558 | 0.00825 | 0.0104 | 0.00932 | 0.0121 |
| | | (st.dev) | (0.00386) | (0.00407) | (0.00545) | (0.00498) | (0.00457) | (0.00477) | (0.00479) | (0.00478) |
| | | median | 0.00146 | 0.00421 | 0.00681 | 0.00406 | 0.00717 | 0.0103 | 0.00845 | 0.0120 |
| | Baseline B4: Voxel / dual | mean | 0.00894 | 0.0185 | 0.0251 | 0.0175 | 0.0237 | 0.0306 | 0.0272 | 0.0338 |
| | | (st.dev) | (0.00767) | (0.00937) | (0.0122) | (0.0119) | (0.00984) | (0.0111) | (0.0111) | (0.00958) |
| | | median | 0.00650 | 0.0167 | 0.0224 | 0.0160 | 0.0230 | 0.0302 | 0.0264 | 0.0341 |
| | Baseline B5: Voxel / single | mean | 0.0105 | 0.0211 | 0.0241 | 0.0186 | 0.0314 | 0.0355 | 0.0335 | 0.0424 |
| | | (st.dev) | (0.00970) | (0.0129) | (0.0126) | (0.0132) | (0.0133) | (0.0131) | (0.0133) | (0.0114) |
| | | median | 0.00641 | 0.0180 | 0.0227 | 0.0165 | 0.0302 | 0.0340 | 0.0322 | 0.0441 |

TABLE 8. PLANNING TIME COST

| Time cost (seconds) | Implementation & GPU | Plan length | | |
|---|---|---|---|---|
| | | 1 | 2 | 3 |
| Main Mesh / dual | TensorFlow 1 / RTX 2080 Ti | 4.76s | 9.11s | 14.7s |
| Baseline B2: Mesh / single | | 1.68s | 2.98s | 4.71s |
| Baseline B4: Voxel / dual | | 88.3s | 192s | 335s |
| Baseline B5: Voxel / single | | 22.2s | 33.9s | 45.9s |
| Main Mesh / Dual | JAX / RTX 3090 | 2.65s | 4.44s | 6.49s |

Planning on ground truths instead of refined VtM estimates (baseline B1) produces slightly better scores for most cases, but the difference is modest. This indicates that the quality of the VtM net's mesh estimates is sufficient for use in this planning system, at least when operating on simulation data.

Dual-net planning outperforms single-net planning (baseline B2), which shows the effectiveness of the epistemic uncertainty avoidance for planning on our incompletely sampled domain. Furthermore we see that the explicit strategy (i.e. inclusion of an explicit uncertainty signal) improves performance compared to the implicit dual-net strategy (baseline B3).

Regardless of the shape representation format used by the planning system, the result of a planning session is a mesh representation generated by the simulation environment. Hence in contrast to predictions, planning outcomes can be compared directly in mesh form between the mesh- and voxel-based versions of the system. We see that the mesh-based system obtains better accuracy when evaluated in mesh format, for both dual- (baseline B4) and single-net (baseline B5) planning. However, this comparison is somewhat unfair in that the voxel-based system is set up to optimise accuracy measured in voxel format, and may outperform the mesh version when evaluated in voxel format. Hence we also convert outcomes to voxel form, and report voxel-based accuracy in Table 7. We see that even with accuracy measured in voxel format, mesh-based planning outperforms voxel-based planning.

Planning time cost is shown in Table 8. Time cost for the main configuration and baselines were measured on a single GeForce RTX 2080 Ti GPU, with the system implemented in TensorFlow (Abadi et al., 2015). Additionally, time cost for a more optimised JAX (Bradbury et al., 2018) implementation of the main configuration was measured on a single GeForce RTX3090 GPU. As is to be expected, planning times are longer for dual-net planning than single-net planning. Calculation of the epistemic uncertainty value $u$ requires decoding of intermediate states, which adds substantial processing cost. However, planning times remain on the order of seconds for the mesh-based system. Note that each run of the planning process considers 12800 to 25600 manipulation sequences, while processing a single manipulation in simulation typically takes 20 to 180 seconds. With regards to the voxel versions, planning times are longer than those reported in (Arnold and Yamazaki, 2019a). Some caveats apply here. For fairness of the accuracy comparison, the number of parallel search strains has been increased compared to (Arnold and Yamazaki, 2019a), to match the mesh version. There is a trade-off between planning accuracy on the one hand and the number of search strains and search iterations on the other hand. The settings used here strike what we deem a good balance for the mesh version, but outside of a baseline comparison, different settings may be more suitable for the voxel version. Additionally, we ran into GPU memory limits when running the dual-net voxel baseline with this same number of parallel strains, so instead of 256 parallel strains, we ran four sequential batches of 64 strains. The long planning times of the voxel version do point to an advantage of the mesh version in terms of computational cost.

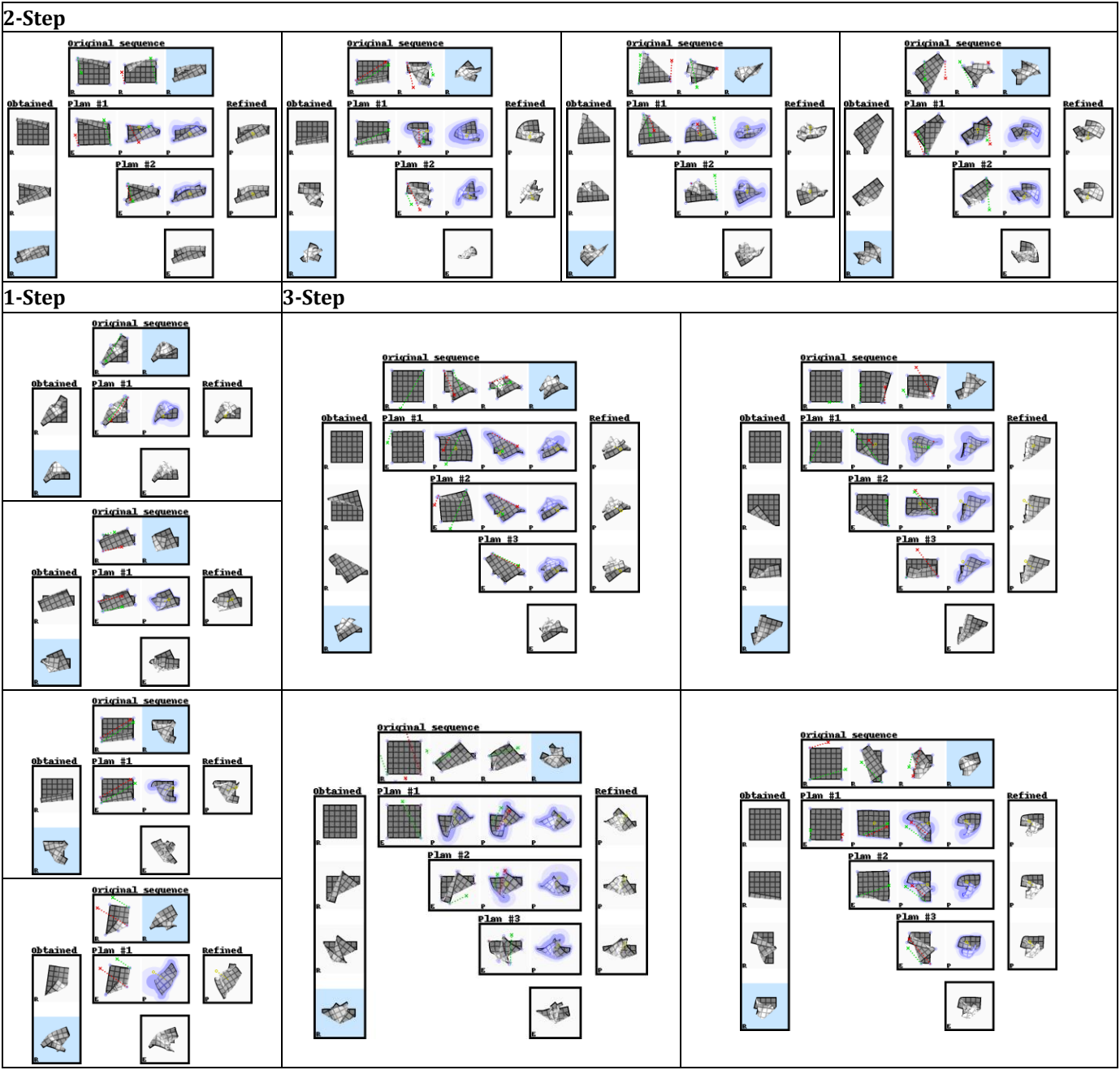

Figure 12. Representative examples of planning & execution sessions (test set data, dual-net planning). In each panel, the top row shows the original manipulation sequence. The system only sees its estimate of the current state and the final state of the original sequence (i.e. the goal state, marked with light blue background). The left column shows the sequence of cloth shapes obtained over the course of the session. The final outcome is marked with a light blue background. Rows marked "Plan #$i$" show the $i^{th}$ plan generated in the session. Each plan starts from an estimate (DMR) of the current shape, generated through the VtM net and refinement. with subsequent shapes being predictions (PMRs) generated by the pEM*D net. Under the last plan, we see the system's estimation of the obtained outcome. The right column shows the result of applying the refinement procedure to the predicted outcome of each plan. These are added for illustration, and not used by the system. They represent a plausible deterministic shape drawn from the probabilistic prediction of each plan's outcome. All shapes are marked in their bottom left corner to indicate the shape type: R = real, E = estimation, P = prediction. Red and green dotted lines show manipulation trajectories. Lilac circles mark grasp point candidates.

## 6.4. FAILURE MODES

The existence of large-error outliers is evident in Figure 11. We observe a few different failure modes. One originates in the grasp point detection routine. Sometimes a corner of the refined VtM estimate is less pronounced than in the ground truth and fails to be detected as graspable. Consequently, plan search fails to consider the grasp point, which can make it impossible to find the optimal manipulation. We also observe errors in the z-ordering of cloth layers in shapes with multiple overlapping folds. For example, when the target shape folds the cloth in two over its x axis and then over its y axis, the generated plan may sometimes fold in reverse order. Also, as mentioned above, the VtM net sometimes estimates incorrect z-ordering, which invites such errors in planning. As the simulated cloth is quite thin, vertex position difference between shapes that differ only in z-ordering turn out small. As such, alternative orderings likely represent local minima in the search space. Loss definitions that accentuate erroneous z-ordering should help resolve this issue.

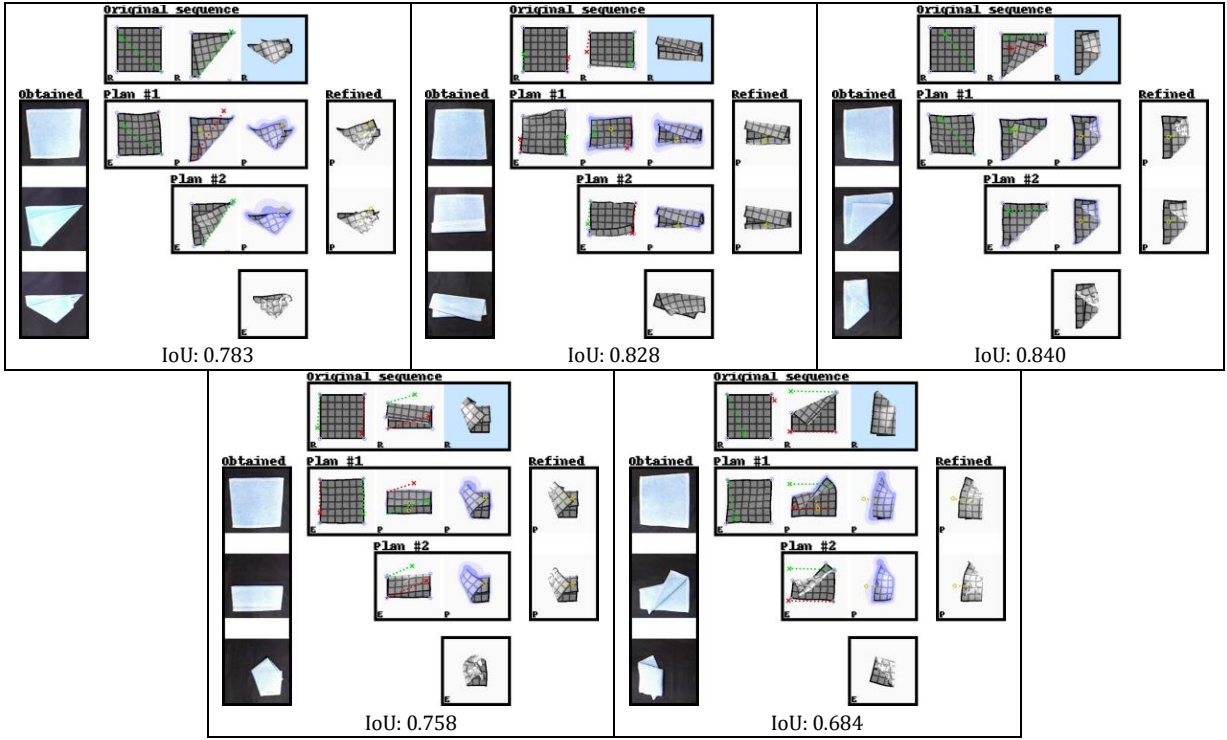

Figure 13. Two-step manipulation sequences planned by the system, performed by a dual-handed robot on real cloth. See Figure 12 for the figure format. The right-most column in each example shows the sequence of actually obtained physical cloth shapes. Input states for plan generation are obtained by shape estimation on voxelised point cloud data of the real cloth. Real cloth shapes are captured at a slight angle due to the camera placement. Plans were rotated by multiples of 90° degrees around the z-axis in order to accommodate limitations of the the robot's range, and images of real cloth shapes are rotated accordingly. Scores are Intersection-over-Union scores over mask images computed for the goal and outcome, indicating the similarity of the top-down silhouettes of goal and outcome shapes, with 1.0 corresponding to a perfect match.

## 6.5. PLANNING (HARDWARE)

We integrated the planning system with a dual-handed robot platform (HIRO, Kawada Robotics), and performed qualitative experiments on real cloth. There are significant differences between the simulated cloth and our real cloth, in particular in fabric thickness, friction with the work surface, and elasticity (stiffness). We manually selected five examples for which we expect the manipulation sequence to produce similar results on our simulated and real cloth. The mesh of the input state is estimated from a voxelised point cloud obtained using a Kinect depth camera mounted on the robot's head, providing a near-top-down perspective on the cloth. We performed interleaved planning and execution following the same procedure as in the simulation experiments. Figure 13 shows results in the same format as Figure 12, and still frames of one example case are shown in Figure 1. We observe that plans closely resembling the original manipulation sequences were generated, and most outcomes clearly resemble the goal states. For many manipulations seen here, close inspection reveals that the trajectories slightly overshoot the optimal release point, leading to folds that are slightly deeper than intended. This is due to a discrepancy in elasticity between the simulated and real cloth. The simulated cloth stretches somewhat under its own weight when lifted, causing it to rebound slightly upon release. The planning system, anticipating this behaviour, places the release points slightly beyond the grasped points' intended final positions. The real cloth, however, stabilises without rebounding, resulting in the slight overshoot seen here. We quantify similarity of the outcomes to the goals using IoU (Intersection-over-Union) scores computed over top-down mask images (i.e. silhouettes) of the goal and outcome. Scores are shown in the figure. Shape estimation accuracy on real cloth is sufficient on the initial and intermediate shapes to allow for effective planning, but estimation on the final outcome is seen to be challenging in cases where many layers of cloth overlap. This is partly due to the difference in thickness between the simulated and real cloth, which causes increasing divergence between the voxel representations of simulated and real cloth as the number of stacked layers increases.

7. Discussion

Our experimental results on simulated cloth confirm that accuracy is significantly improved by the use of mesh representations and dual-net planning with an explicit epistemic uncertainty signal. The advantage of mesh representation holds up even when we evaluate outcomes in voxel form. This likely reflects the fact that accuracy in mesh format implies accuracy in voxel format, but not vice-versa.

The present work deviates from the trend of training systems to encode raw sensor data (pixels/voxels) into latent representations (Arnold and Yamazaki, 2019a; Hoque et al., 2020; Wahlström et al., 2015; Yan et al., 2020). Central to this trend is the idea that the deep learning training procedure will discover the data's deep structure and encode it into a latent representation format optimised to the task at hand. Our results suggest that this may be optimistic. The mesh and voxel versions of the system operate on identical input states (voxel representations), and internally generate latent representations of the same dimensionality. If the training procedure were maximally effective at discovering the data's deep structure, we would expect both versions to produce similar and equally effective latent representations. The performance gap thus suggests that the training procedure (a fairly typical deep learning training procedure) is not actually capable of discovering the data's deep structure all by itself.

The advantage of setting goal states in mesh format bears emphasising. When we consider tasks such as garment folding or knot-tying, we aim not just for visual similarity from a given viewpoint, but for structural similarity to the goal state. This is hard to achieve when the system does not represent the object's structure in a form that allows for comparison to a to structured goal representation.

With regard to time cost, dual-net planning is more demanding than single-net planning. This is in large part due to the need to decode the predicted intermediate and final states from latent form to full form, in order to allow their comparison between nets. However, the switch to comparatively lightweight mesh representations offsets the additional cost. On current high-end consumer-grade GPUs, planning times on the order of a few seconds are retained.

We find that in numerous cases, planning outcomes more closely resemble the goal state than the system's own predictions for the generated plan. This may seem counterintuitive, but should not be surprising given the system's planning logic. In principle, close resemblance between the goal state and the prediction for the optimal action sequence is no requirement for planning to find the optimal action sequence. What is required is merely that the prediction for the optimal action sequence better approximates the goal than the predictions for non-optimal action sequences. This creates significant leeway in terms of prediction accuracy. The use of explicit aleatoric uncertainty in the mesh format helps the system exploit this leeway. For parts of the state that the system cannot predict with precision, it indicates lack of confidence with high $\sigma$ values. High $\sigma$ values reduce the impact of these parts on the planning loss (recall that we use a likelihood loss). Parts that are predicted with relatively high confidence thus dominate plan search. When the relatively predictable parts of the deformation process sufficiently constrain action search, near-optimal actions can be found despite significant uncertainty in other parts of the prediction. If the action is correct and the deformation dynamics are consistent, the hard-to-predict parts will fall in place when the action is executed, even if the system could not predict that they would. Hence the combination of explicit aleatoric uncertainty and backpropagation-based planning with a likelihood loss allow for effective planning even in presence of substantial prediction uncertainty.

As will be clear from the result images in Section 6, our dataset includes complex shapes that would have limited use in a practical household support setting. A simpler dataset may well suffice to obtain sufficient manipulation skills for practical tasks. However, while the shape repertoire may be somewhat excessive, the choice not to constrain the data repertoire to a set of basic, common cases is fundamental to our goal. Recall that we pursue a generalised affinity for cloth manipulation. Pursuing generality implies that the system can generate plans for many goals that have no practical use. This may seem inefficient from a pragmatic point of view, but consider the generality of human planning abilities. Humans can generate plans for obtaining countless outcomes that we would never have any use for, as well as valuable solutions for novel situations. The alternative to pursuing a high degree of generality is to assume that we can imagine and account for all the situations a robotic system will find itself in, which may be a strong assumption to make when a robot is supposed to function flexibly in an uncontrolled environment.

8. Conclusions & Future work

We presented a system for cloth manipulation planning, building on the system proposed in (Arnold and Yamazaki, 2019a). We adopted mesh representations to eliminate much of the ambiguity that hampers voxel-based planning. The problem of obtaining mesh representations was addressed by introducing a neural network-based shape estimation routine into the system. The introduction of mesh representations was demonstrated to significantly improve accuracy for both prediction and planning. We also addressed the problem of planning with incomplete domain knowledge. Having to sample the full space is both costly and wasteful (many possible manipulations are of no practical value), limiting applicability to real-world settings. This requirement was eliminated here by adopting a dual-network planning strategy that explicitly quantifies epistemic uncertainty, and avoids manipulations for which confidence is low. We demonstrated the effectiveness of this approach, and showed that the use of an explicit epistemic uncertainty signal is more effective than the implicit strategy that merely combines the search losses from the two nets. Finally, the need to work with discrete sets of graspable points was addressed by incorporating grasp point detection into the system. These improvements make significant strides towards practical applicability.

Nonetheless, a number of open issues and avenues for further development remain. In the present system, planning is closed-loop at the granularity of full manipulation steps: after executing a manipulation, the system observes its result, and replans accordingly. Execution of the individual manipulation steps, however, occurs without feedback. Correction of trajectories mid-manipulation can be achieved by applying many of the same concepts we used for planning (shape estimation, prediction, manipulation generation through backpropagation) at a finer timescale. We pursue online trajectory correction based on these ideas in parallel work (Tanaka et al., 2021). Eventually we aim to integrate the two levels of time granularity into a single system, to combine multi-step manipulation planning with fine-grained closed-loop control of individual manipulation steps.

We have focused on simple square cloth as the object of manipulation. Expanding the system to other cloth topologies is straightforward in theory. No part of the system relies on the square topology. A convenient feature of encoder and decoder architectures with full connectivity is that they can be used for any topology by simply matching the number of input units to the number of vertices in the mesh representation. The only aspect of the system that would require significant modification is the handling of Cartesian equivalence (see Section 4.6), as this is a topology-specific property. For more complex topologies there will be fewer or no equivalent mesh representations of a given shape, allowing for some system simplification. Actual system performance on alternative topologies, however, remains to be assessed. We intend to experiment with common clothing topologies such as t-shirts and trousers in future work. Note that for many non-rectangular topologies, not all geodesic coordinates would correspond to valid positions on the cloth, meaning that there would be invalid regions in the action space. We expect our epistemic uncertainty avoidance strategy to be effective in avoiding these regions, but this too remains to be confirmed.

Discrepancies in material properties between the simulated and real cloth are an obstacle to real-world generalisation, as we observed in our hardware experiments. Matching the simulated cloth more closely to the real cloth would improve performance, but this strategy is of limited practical value, as material properties vary substantially between different real cloth objects as well. A structured method for handling material-induced uncertainty is currently under development.

Our planning evaluation experiments assume plan length to be given. This is not a realistic assumption. Also, in practical applications it would also be sensible to allow for some leeway in the number of manipulations, because in general accuracy will take precedence over minimising the number of manipulations performed. Planning with variable plan lengths can be achieved using a planning loss that compares predicted intermediate states against the goal state.

Whereas mesh representation avoids the ambiguities of voxel representations in the planning process, input to our system's shape estimation routines remains in voxel format, and here challenges due to voxel representation ambiguity remain. One strategy that could help reduce ambiguity is to eliminate voxelisation altogether, and operate on pointcloud data directly. Voxel representations are particularly easy to process using NNs, but recent years have seen advances in direct NN processing of pointcloud data as well, in the invention of PointNet (Qi et al., 2017) and subsequent architectures building thereon. We hypothesise that performing mesh estimation on pointcloud data without the intervening conversion to voxels has the potential to reduce (though not eliminate) ambiguity and produce better estimations, by picking up on finer disambiguating details that are lost in voxelisation. We are experimenting with PointNet-style estimation nets, but our best attempts so far fall short of our VtM net by a small margin.

In the present paper we mostly treated shape estimation and plan generation as separate systems, implemented as separate networks trained separately. Tighter integration is possible, and may be beneficial. We consider a few possibilities. In the present work we trained the pEM*D net on deterministic ground truth meshes, but it could be modified to take probabilistic VtM estimates as input directly. We brought an element of expectation into shape estimation by initialising refinement with a prediction of the present state (when available), but the VtM net could be extended to take a prediction as a secondary input, to allow expectations to reduce ambiguity in the probabilistic shape estimation stage. It should also be possible to integrate the VtM net's functionality into encoder module E. However, some of these modifications would complicate integration of prior topological knowledge about the cloth into the system, as we presently do through refinement.

Finally, we used a memory trace to allow past states and manipulations to influence plan generation, but its implementation is rudimentary, and its role has not been studied in detail. We believe the memory trace would be important in tasks with prominent occlusion, and aim to experiment with such tasks in future work.

ACKNOWLEDGEMENTS

This work is partly supported by NEDO and JSPS KAKENHI.

CONFLICT OF INTEREST

The Authors declare that there is no conflict of interest.